\documentclass[sigconf]{acmart}
\usepackage{multirow}
\usepackage{booktabs}
\usepackage{threeparttable}
\usepackage{algorithm}
\usepackage{enumitem}
\usepackage{algorithmic}
\usepackage{amsthm,amsmath,mathrsfs}
\usepackage{subfigure}
\usepackage{tikz}
\usepackage{xcolor}
\newcommand{\tpm}[1]{\tiny{$\pm\Delta$#1}}
\AtBeginDocument{%
  \providecommand\BibTeX{{%
    \normalfont B\kern-0.5em{\scshape i\kern-0.25em b}\kern-0.8em\TeX}}}

\copyrightyear{2023}
\acmYear{2023}
\setcopyright{rightsretained}
\acmConference[KDD '23] {Proceedings of the 29th ACM SIGKDD Conference on Knowledge Discovery and Data Mining}{August 6--10, 2023}{Long Beach, CA, USA.}
\acmBooktitle{Proceedings of the 29th ACM SIGKDD Conference on Knowledge Discovery and Data Mining (KDD '23), August 6--10, 2023, Long Beach, CA, USA}
\acmPrice{}
\acmISBN{979-8-4007-0103-0/23/08}
\acmDOI{10.1145/3580305.3599372}
\begin{document}

\title{Graph Neural Processes for Spatio-Temporal Extrapolation}

\author{Junfeng Hu}
\affiliation{%
  \institution{National University of Singapore}
  % \country{Singapore}
}
\email{junfengh@u.nus.edu}
\author{Yuxuan Liang}
\authornote{Y. Liang is the corresponding author of this paper.}
\affiliation{%
  \institution{Hong Kong University of Science and Technology (Guangzhou)}
  % \country{Singapore}
}
\email{yuxuanliang@hkust-gz.edu.cn}
\author{Zhencheng Fan}
\affiliation{%
  \institution{University of Technology Sydney}
  % \country{Australia}
}
\email{zhencheng.fan@student.uts.edu.au}
\author{Hongyang Chen}
\affiliation{%
  \institution{Zhejiang Lab}
  % \country{Singapore}
}
\email{dr.h.chen@ieee.org}
\author{Yu Zheng}
\affiliation{%
  % \institution{JD Intelligent Cities Research}
  \institution{JD Intelligent Cities Research \& JD iCity, JD Technology}
  % \country{China}
}
\email{msyuzheng@outlook.com}
\author{Roger Zimmermann}
\affiliation{%
  \institution{National University of Singapore}
  % \country{Singapore}
}
\email{rogerz@comp.nus.edu.sg}
% \author{Junfeng Hu}
% \email{junfengh@u.nus.edu}
% \affiliation{%
%   \institution{National University of Singapore}
% }

% \affiliation{%
\renewcommand{\shortauthors}{Junfeng Hu et al.}

\begin{abstract}
We study the task of spatio-temporal extrapolation that generates data at target locations from surrounding contexts in a graph. 
This task is crucial as sensors that collect data are sparsely deployed, resulting in a lack of fine-grained information due to high deployment and maintenance costs. 
Existing methods either use learning-based models like Neural Networks or statistical approaches like Gaussian Processes for this task. However, the former lacks uncertainty estimates and the latter fails to capture complex spatial and temporal correlations effectively.
To address these issues, we propose Spatio-Temporal Graph Neural Processes (STGNP), a neural latent variable model which commands these capabilities simultaneously. Specifically, we first learn deterministic spatio-temporal representations by stacking layers of causal convolutions and cross-set graph neural networks.
Then, we learn latent variables for target locations through vertical latent state transitions along layers and obtain extrapolations. 
Importantly during the transitions, we propose Graph Bayesian Aggregation (GBA), a Bayesian graph aggregator that aggregates contexts considering uncertainties in context data and graph structure.
Extensive experiments show that STGNP has desirable properties such as uncertainty estimates and strong learning capabilities, and achieves state-of-the-art results by a clear margin. %Our implementation is anonymously available at \url{https://anonymous.4open.science/r/STGNP_Anonymous-53FB}.
\end{abstract}

\begin{CCSXML}
<ccs2012>
   <concept>
       <concept_id>10002951.10003227.10003236</concept_id>
       <concept_desc>Information systems~Spatial-temporal systems</concept_desc>
       <concept_significance>300</concept_significance>
       </concept>
 </ccs2012>
\end{CCSXML}

\ccsdesc[300]{Information systems~Spatial-temporal systems}


\keywords{Spatio-temporal data mining, neural processes, data extrapolation}



\maketitle

\section{Introduction}
\begin{figure}
\setlength{\belowcaptionskip}{-4pt}
  \centering
  \includegraphics[width=0.95 \linewidth]{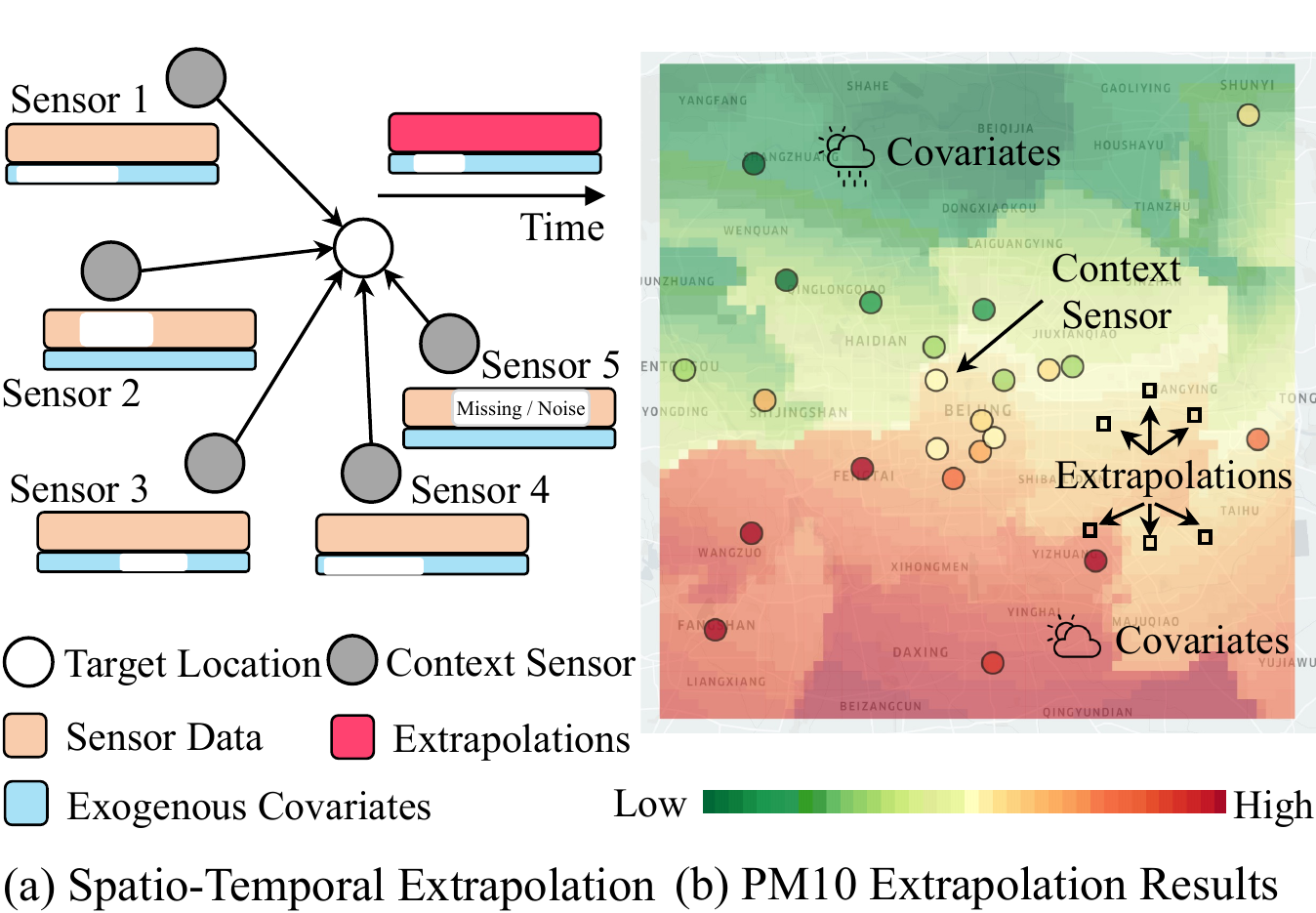}
  \caption{(a) Context sensors 1-5 are utilized to generate data of a target location, considering exogenous covariates and the graph structure. (b) Extrapolations of our STGNP.}
    \label{figure_intro}
\end{figure}

Spatio-temporal graph data, such as air quality readings \cite{zheng2015forecasting,liang2022airformer}, traffic flow data \cite{wu2019graph,liang2019urbanfm,pan2019urban}, and climate data \cite{lin2022conditional} reported by deployed sensors,  are ubiquitous in the physical world. Analyzing such data fosters a variety of applications for smart cities, enhancing people's lives and helping decision-making \cite{jin2023spatio}.  Ideally, the data should be fine-grained to realize its benefits but it is often impractical to deploy and maintain sufficient sensors in an area of interest because of the high expenditure \cite{shao2021trajforesee,wang2022inferring}. For example, a professional station to measure air quality can cost around \$200,000 to build and \$30,000 per year to maintain~\cite{zheng2013u}. As a result, many applications have to rely on sparse data, leading to suboptimal solutions. Thus, finding ways to approximate the data in areas with no sensors has become a pressing issue.

In this paper, we study the problem of \textit{spatio-temporal extrapolation}, which involves estimating a function that predicts the spatio-temporal data at target locations of interest areas based on the surrounding context nodes and related exogenous covariates, operated in a graph structure, as shown in Figure~\ref{figure_intro}(a).  
As an example, we use air quality extrapolation \cite{zheng2013u} illustrated in Figure~\ref{figure_intro}(b). We utilize the air quality index from context sensors to extrapolate the indexes at the target locations, taking into account covariates such as weather conditions that can also impact air quality.
%we assume $Y^*$ is the air quality index affected by factor weather conditions $E^*$ (e.g., sprinkle, cloudy). The function leverages $({\mathcal{X}_\mathcal{C}}, {\mathcal{Y}_\mathcal{C}})$ of surrounding sensors to extrapolate $Y^*$ at the target location. 

To achieve our goal, one pivotal property that needs to be considered is spatio-temporal correlations, i.e., the spatial dependencies within the graph and the temporal dependencies along the time axis. To capture these correlations, Neural Networks (NNs), especially Spatio-Temporal Graph Neural Networks (STGNNs), nowadays have become a favorite due to their tempting learning competence~\cite{hanfine,wu2019graph}. 
However, NNs have two main limitations: (i) \emph{They lack the sought-after ability to estimate uncertainties.} Sensors often produce signals with different levels of ambiguity, such as noisy or missing observations due to unpredictable factors such as network disruptions. Explicitly capturing these uncertainties has proved to be a boon for making reliable decisions~\cite{wang2019deep,wen2023diffstg}. However, most NNs are deterministic and unable to account for these uncertainties.  
(ii) \emph{Their ability to generalize to new data is limited.}
NNs require a large amount of training data to parameterize a function. Nevertheless, their reliance on parameters hinders their adaptability in unpredictable real-world environments, as the model needs to be retrained whenever the environment changes. Additionally, their sensitivity to hyperparameters demands a hyperparameter search to attain optimal performance.

The limitations of NNs have led researchers to draw inspiration from probability models, with Gaussian Processes (GPs) being one potential approach~\cite{seeger2004gaussian}. 
GPs define a stochastic process in which the spatio-temporal relations are modeled by various kernels~\cite{patel2022accurate}.
Their Bayesian principle and non-parametric nature enable them to handle uncertainty and generalize well to a wide range of functions~\cite{luttinen2012efficient}.
However, GPs suffer from limited expressivity of kernels, which can be disadvantageous.
To address these issues, Neural Processes (NPs)~\cite{garnelo2018neural} have emerged as a promising approach. NPs learn to construct stochastic processes by parameterizing neural networks in which an aggregator is introduced to integrate contexts. By combining the strengths of both NNs and GPs, NPs offer a compelling framework for spatio-temporal extrapolation.

Unfortunately, NPs cannot be applied directly to spatio-temporal graph data due to the following factors: (i) \emph{They struggle to learn temporal dependencies effectively.} Existing NPs ~\cite{singh2019sequential,qin2019recurrent} utilize latent state transitions to capture temporal relations recurrently. However, transitions tend to only focus on learned latent variables, disregarding the contextual information at later recurrent steps.
This phenomenon, known as transition collapse, can impede learning over long sequences~\cite{singh2019sequential}.
(ii) \emph{They are incapable of modeling spatial relationships in a graph.} Existing NPs' aggregation operations~\cite{gordon2019convolutional,kim2019attentive,volpp2020bayesian} lack the ability to model complex spatial relations defined in a graph. In addition, their deterministic nature makes them suboptimal for aggregating data with varying levels of ambiguity, such as missing values and noise in Figure~\ref{figure_intro}(a).

To tackle these challenges, we propose Spatio-Temporal Graph Neural Processes (STGNP) for spatio-temporal extrapolation over graphs. 
STGNP has two stages: the first stage leverages a deterministic network to learn spatio-temporal representations of nodes.
Instead of relying on a recurrent structure, the temporal dynamics are modeled by stacking convolution layers in a bottom-up way~\cite{aksan2019stcn}, while the spatial relations are captured by cross-set graph neural networks. 
In the second stage, we employ state transitions to aggregate latent variables of target nodes following a top-down manner. Here, the transition assumes horizontal time independence and incorporates long-range temporal evolution from the upper layers. As the number of transitions only relates to the number of stacked layers, much smaller than the sequential length, our model naturally exhibits resistance to transition collapse.

For the aggregator in each transition, we claim that different context nodes have different levels of importance. Motivated by~\cite{volpp2020bayesian}, we propose Graph Bayesian Aggregation (GBA) that directly aggregates distributions over latent variables regulated by the graph. Intuitively, a context node contributes less to the latent distribution if it is far from the target location or exhibits a high degree of ambiguity recognized by the module. This design explicitly considers graph structure into NP's aggregator and enhances the model's capability to capture node uncertainties.
% This design explicitly enhances the model's capability to capture node uncertainties and improve the overall performance.

In summary, our main contributions are summarized as follows:
\begin{itemize}[leftmargin=*]
    \item We propose Spatio-Temporal Graph Neural Processes. To the best of our knowledge, this is the first work that generalizes NPs to spatio-temporal graph modeling. 
    STGNP captures uncertainties explicitly and can generalize to different functions robustly, which is a major advantage over NNs approaches. 
    Additionally, STGNP is able to learn temporal relations and graph data effectively, which sets it apart from classical NPs models.
    \item We introduce Graph Bayesian Aggregation, a Bayesian method for aggregating context nodes, which allows the aggregator to model graph structure and uncertainties for context nodes.
    \item We conduct comprehensive experiments to evaluate the performance and properties of STGNP. Our results demonstrate that STGNP outperforms state-of-the-art baselines by a significant margin and exhibits compelling properties, such as uncertainty estimates, high generalizability, and robustness to noisy data.
\end{itemize}

\section{Preliminaries}
We first define concepts and notations of spatial-temporal data on graphs. Then, we introduce the basic concepts of neural processes.
%We first describe the basic concepts of neural processes. Then, we define notations and spatial-temporal data over graphs and formalize the problem of data extrapolation.
\subsection{Definitions and Notations}
\paragraph{Definition 1 (Graph)} We represent nodes as a graph $\mathcal{G}=(\mathcal{V}, \mathcal{E})$, where $\mathcal{V}$ is the vertex set and $\mathcal{E}$ is the edge set defining the weights between nodes. The $K$-hop neighborhood of a node $v\in \mathcal{V}$ denoted by $\mathcal{N}_k(v)$ is the set of nodes that are reachable from $v$ with $K$ steps. Based on $\mathcal{E}$ and $K$, an $K$-hop adjacency matrix $A^K$ is derived to measure the non-Euclidean distances between connected neighbors.
\paragraph{Definition 2 (Spatio-Temporal Data)} Observed signals are retrieved from each node in the graph. We use $Y_i=(y_{i,1}, .., y_{i,t}, .., y_{i, T}) \in \mathbb{R}^{T\times d_y}$ to denote data of node $i$ that is measured over a time window $T$, where $d_y$ is the number of features. $Y=(Y_1, .., Y_n, .., Y_N) \in \mathbb{R}^{N\times T \times d_y}$ is denoted as a signal tensor of all nodes over the window $T$, where $N$ is the total number of observable nodes in the graph.
\paragraph{Definition 3 (Exogenous Covariates)} Exogenous covariates benefit the learning process as they usually have notable correlations with node data. These covariates are readily available from different sources. For instance, weather conditions can affect air pollutant data and they are collected from weather stations. We denote these factors as a tensor $X\in \mathbb{R}^{N \times T \times d_x}$ and consider them explicitly.  

\subsection{Neural Processes}
Neural Processes~\cite{garnelo2018neural} construct stochastic processes that map $x\in \mathbb{R}^{d_x}$ in an input domain to $y\in \mathbb{R}^{d_y}$ in an output space, conditioned on a context set $\mathcal{C}=\{(x_n, y_n)\}_{n=1}^N$ of observed input-output pairs. NPs follow the same principle as Gaussian Processes except the stochastic process is learned implicitly by neural networks. 
Specifically, NPs define a conditional latent variable framework, where the distribution of a latent variable $z$ is described by a learned conditional prior $p(z| \mathcal{C})$ from the context set. Then, with inputs of target variables $X_\mathcal{D}=\{x_m\}_{m=1}^M$ in a target set $\mathcal{D}$, a likelihood module $p({Y}_\mathcal{D} | {X}_\mathcal{D}, z)$ is trained to predict the corresponding output variables $Y_\mathcal{D}$. The following posterior predictive likelihood formulates the generative process of NPs:
\begin{equation}
    p({Y}_\mathcal{D} | {X}_\mathcal{D}, \mathcal{C}) = \int p({Y}_\mathcal{D} | {X}_\mathcal{D}, z)p(z | \mathcal{C}) dz.
\label{NP_gen}
\end{equation}
In practice, NPs assume the target variables are independent, decomposing the likelihood such that $p({Y}_\mathcal{D}| {X}_\mathcal{D}, z)$ is factorized as $\prod_{m=1}^M p(y_m | x_m, z)$, where $M=|\mathcal{D}|$. NPs organize a meta-learning framework where each pair of $\{\mathcal{C}, \mathcal{D}\}$ constructs its own stochastic process, making it less parametric dependent whit strong generalizability. Note that conditions $\mathcal{C}$ in the prior should be aggregated by a permutation-invariant function (e.g., mean, attention) to define a stochastic process, according to Kolmogorov Extension Theorem~\cite{oksendal2003stochastic}.
As the marginalization of the latent variable $z$ is normally intractable, the model is usually trained either by Monte-Carlo (MC) sampling to estimate Equation~\ref{NP_gen} directly~\cite{foong2020meta} or by variational approximation of maximizing the evidence lower bound (ELBO)~\cite{garnelo2018neural}:
\begin{align}
    \log p({Y}_\mathcal{D} | {X}_\mathcal{D}, \mathcal{C}) \geq \mathbb{E}_{q(z|\mathcal{C}\cup \mathcal{D})} \left[ \sum_{m=1}^m \log \frac{p(y_m|x_m,z)p(z|\mathcal{C})}{q(z|\mathcal{C}\cup \mathcal{D})} \right],
\end{align}
where $q(z|\mathcal{C}\cup \mathcal{D})$ and $p(y_m|x_m,z)$ are the approximated posterior and the likelihood learned by neural networks. As the true conditional prior $p(z|\mathcal{C})$ in the numerator is also intractable, the same module $q(\cdot)$ is employed to approximate $p(z|\mathcal{C}) \approx q(z|\mathcal{C})$.

\section{Methodology}
\begin{figure}
  \centering
  \includegraphics[width=1.\linewidth]{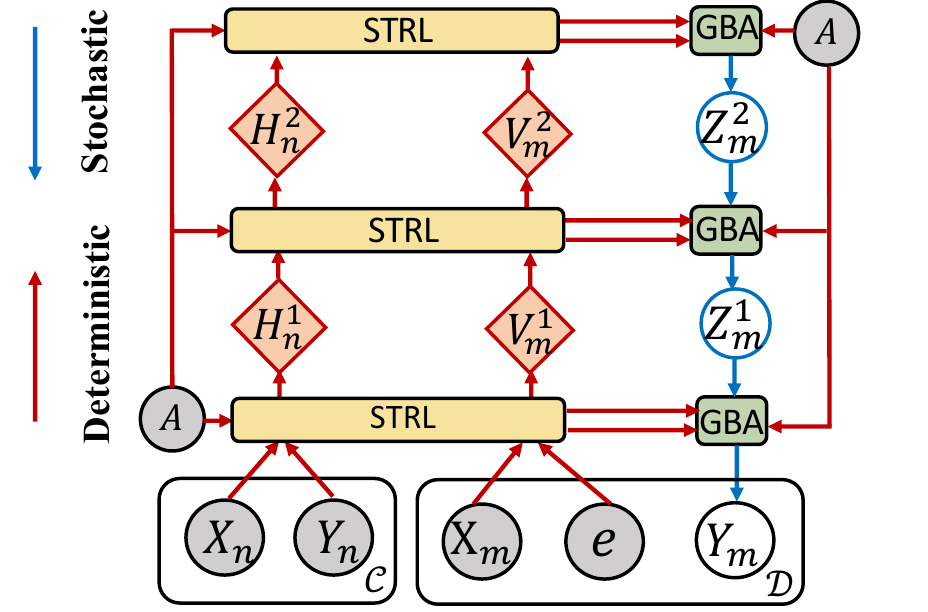}
  \caption{Graphical model with three layers for illustration. $V_m^l, Z_m^l$, $H_n^l\in \mathbb{R}^{T\times d_l}$ are deterministic representations, latent variables of a target node $m$ and representations of a context node. $e\in \mathbb{R}^{d_0}$ is the learnable target token. The shadow circle, STRL, GBA denote an observed variable, a spatio-temporal representation learning, and Graph Bayesian Aggregation.}
    \label{figure_overall}
\end{figure}

%where Figure~\ref{figure_overall} illustrates its graphical model.
In this section, we propose STGNP, a neural latent variable model to enhance spatio-temporal extrapolation. As its graphical model illustrates in Figure~\ref{figure_overall}, the key pipeline is to learn deterministic representations (STRL) and stochastic latent variables (GBA) in two stages. We first introduce the problem of spatio-temporal extrapolation. Then, we describe the deterministic stage for learning spatio-temporal representations and derive Graph Bayesian Aggregation to aggregate contexts in the stochastic stage. Finally, we introduce the generative process and the optimization procedure.
Note that as target nodes are independent, we only discuss a single target node $m$ in the following sections for brevity. 

\subsection{Problem Statement} We formulate spatio-temporal extrapolation in the NPs framework and first define the context set $\mathcal{C}$ containing nodes with exogenous covariates and observed data $\{(X_n, Y_n)\}_{n=1}^N \in \mathbb{R}^{N\times T \times (d_x+d_y)}$. Our goal is to learn a posterior predictive distribution $p(Y_\mathcal{D}|X_\mathcal{D}, \mathcal{C}, A)$ to generate $Y_\mathcal{D} \in \mathbb{R}^{M \times T \times d_y}$ over the same time period in the target set $D$ where $M$ is the total number of target nodes, given the covariates $X_\mathcal{D}$, the context set $\mathcal{C}$, and the adjacency matrix $A$. In this paper, we use subscript $m$ and $n$ to index target and context nodes respectively, and adopt the terms location, node, and sensor interchangeably.

\subsection{Spatio-Temporal Representation Learning}

The deterministic stage has three building blocks to capture spatial and temporal correlations: learnable target token, dilated causal convolution, and cross-set graph convolution. We introduce them individually and then illustrate the overall learning framework.

\paragraph{Learnable Target Token} Our network takes sensor data and covariates as inputs; however, the data $Y_m$ of a target node is unknown. Existing methods typically preprocess it by either filling it with zeros~\cite{appleby2020kriging,wu2021inductive} or employing linear interpolation to estimate its values~\cite{hu2021decoupling}. However, using zero to represent target variables can be confusing, as it may be interpreted as the semantic zero within the dataset. Moreover, interpolation incurs large errors, which also hampers model performance. Inspired by Masked AutoEncoder~\cite{he2022masked}, we leverage a shared learnable token $e\in \mathbb{R}^{d_0}$ as embeddings for target nodes, while using an embedding layer with the parameter $W\in\mathbb{R}^{d_y\times d_0}$ to get embeddings for context nodes. The token is optimized by the network to identify a position in the feature space that represents the target node, which avoids inferior preprocessing.

\paragraph{Cross-Set Graph Convolution Layer} Graph convolution is a seminal operation to learn spatial relations in a graph structure. Existing GCNs methods typically treat dependencies over all nodes equally~\cite{wu2021inductive,hu2021decoupling}. However, in our task, relations across the target and context sets take priority due to their direct influence on the target nodes. Based on this insight, we argue that disregarding internal relations within the two sets does not adversely affect performance and propose cross-set graph convolution (CSGCN), in which only dependencies across the set $\mathcal{C}$ and $\mathcal{D}$ are captured.
Specifically, given the representation of the target node  $V_m^{l-1}\in \mathbb{R}^{T\times d_{l-1}}$ at layer $l-1$, we update it by its neighbors $H_n^{l-1}$ in the context set up to $K$-hop, with each neighbor weighted by the adjacency weight $A^k_{m,n}$:
\begin{equation}
    V_m^l = \sum_{k=0}^K \frac{V_m^{l-1} + \sum_{n\in \mathcal{N}^c_k(m)}A^k_{m,n} H_n^{l-1}}{1 + \sum_{n\in \mathcal{N}^c_k(m)}A^k_{m,n}} W^l_k,
\end{equation}
where $W^l_k\in \mathbb{R}^{d_{l-1}\times d_l}$ are learnable parameters and $\mathcal{N}^c_k(m)$ is $k$-hop neighbors of the target node $m$ indexed from $A^k$. Note that when $l=0$, $V_m^0$ is the broadcast target token and $H_n^0$ is the context embeddings. Compared to traditional GCNs, CSGCN offers improved efficiency, reducing the computational complexity from $\mathcal{O}((N+M)^2)$ to $\mathcal{O}(N\times M)$. Despite this efficiency gain, it maintains strong learning capabilities as demonstrated in the experiments.

\paragraph{Dilated Causal Convolution Layer} We employ dilated causal convolutions (DCConv)~\cite{yu2016multi} to capture temporal dependencies. Unlike the recurrent structure, it learns temporal relations over long sequences by stacking causal layers. This approach proves advantageous as the number of layers is considerably smaller than the length of the sequence, mitigating the issue of transition collapse in the later stage. Specifically, at time $t$, a 1D causal convolution learns a temporal representation $h_{i,t}^l\in \mathbb{R}^{d_l}$ for node $i$:
\begin{equation}
    h_{i,t}^l = H_i^{l-1} \star \mathcal{K}^l(t) = \sum_{s=0}^{k-1} \mathcal{K}^l(s) \odot H_i^{l-1}(t-\eta\times s),
\end{equation}
where $H_i^{l-1}\in \mathbb{R}^{T\times d_{l-1}}$ is a node representation at the previous layer,
$\star \mathcal{K}^l$ means a DCConv with the kernel size $c\times d_{l-1}\times d_l$, and $\odot$ is the Hadamard product. The dilation factor $\eta$ is initialized to $1$ with an exponentially increasing rate of $2$~\cite{oord2016wavenet} and zero-padding is used to ensure the inputs and outputs have the same time length $T$.

%Thus, the final receptive field of the top layer $L$ reaches
%This means that the receptive field increases exponentially w.r.t the number of layers, which allows the model to reach the same receptive field with fewer layers. 

\begin{figure}
  \centering
  \includegraphics[width=1.\linewidth]{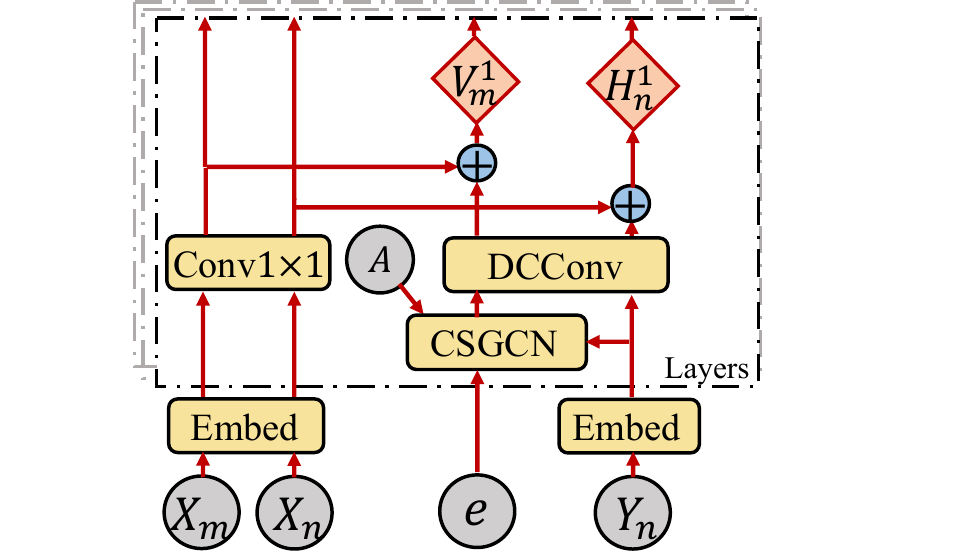}
  \caption{The pipeline of the spatio-temporal representation learning network, where we first capture temporal dependencies using the DCConv and then learn spatial relations by CSGCN. Embed denotes an embedding layer.}
    \label{figure_STRB}
\end{figure}

\paragraph{Learning Framework} As shown in Figure~\ref{figure_STRB}, each layer of the network first applies a CSGCN to model spatial relations, followed by a DCConv to capture temporal dependencies for node representations. Additionally, the features of covariates, learned through a $1\times 1$ convolution, are incorporated into the node representations. Note that we do not explicitly involve covariates in CSGCNs and DCConvs, as they may exhibit different spatio-temporal dependencies or even lack relations in certain scenarios~\cite{tashiro2021csdi}.
By stacking layers with skip connections, the representations of the target node are obtained in which each layer maintains temporal dependencies at various scales, with upper layers capturing long-range relations and lower layers comprising fine-grained information.
Thus, the stochastic stage is able to access different scales through its hierarchical dependency.

\subsection{Graph Bayesian Aggregation}
\begin{figure}[b]
  \centering
  \includegraphics[width=1.10\linewidth]{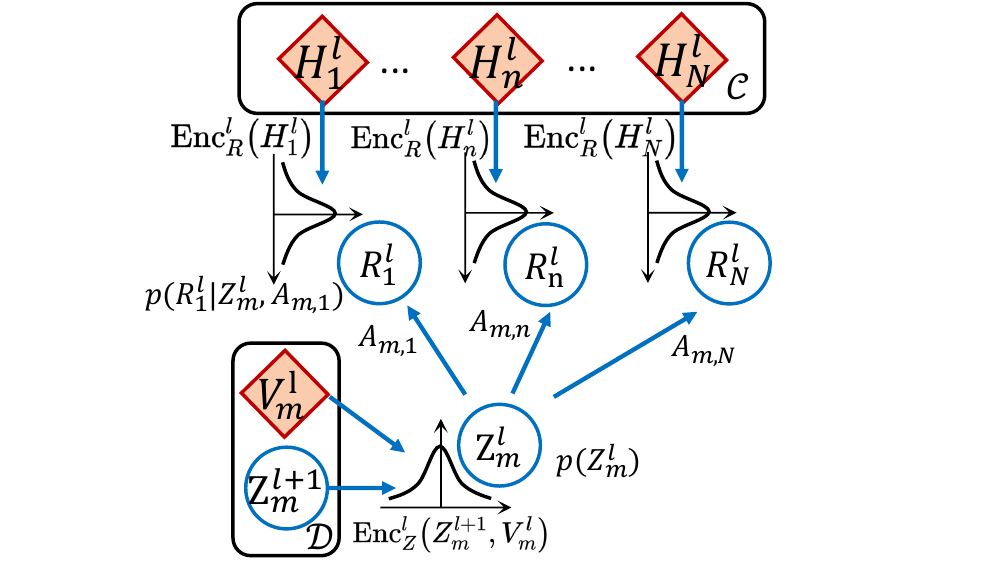}
  \caption{Graph Bayesian Aggregation. $\operatorname{Enc}_Z^l(\cdot)$ and $\operatorname{Enc}_R^l(\cdot)$ are neural networks that learn mean and variance for the prior and latent observation distribution.}
    \label{figure_GBA}
\end{figure}
The core component for the stochastic stage is our proposed Graph Bayesian Aggregation, which aggregates information from context nodes and derives latent variables $Z_m^l\in \mathbb{R}^{T\times d_l}$ describing stochastic processes over target nodes.
%In each layer of the stochastic stage, we propose a Graph Bayesian Aggregation to aggregate the target node's latent variable $z_t$ from context nodes, 
Figure~\ref{figure_GBA} illustrates the aggregation process.
%As the process is independent and identical for a target node at different times, we omit the subscript $t$ to avoid clutter. 
Based on Bayes’ theorem~\cite{bishop2006pattern}, we assume a prior $p(Z_m^l)$ over the target node. Then for each context node $n$, a latent observation model $p(R^l_{n} | Z^l_m, A_{m,n})$ is derived in which its mean conditions on a linear transformation of $Z_m$ and $A_{m,n}$.
Thus once observe $R^l_{n}$, the latent variable $Z^l$ is updated through its posterior:
\begin{equation}
p(Z_m^l|\{(R^l_n, {A}_{m,n})\}_{n=1}^N) = \frac{\prod_{n\in \mathcal{N}^c_1(m)} p(R^l_n | Z_m^l,{A}_{m,n}) p(Z_m^l)}{\prod_{n\in \mathcal{N}^c_1(m)} p(R^l_n)},
\label{posterior}
\end{equation}
where we suppose the latent observations are independent and only consider the $1$-hop neighbor to simplify the computation. The prior $p(Z_m^l)$ follows a factorized Gaussian:
\begin{align}
\begin{split}
    &p(Z_m^l) = \mathcal{N}(Z_m^l | \mu_{Z_m^l}, \operatorname{diag}(\sigma_{Z_m^l}^2)),\\
    &(\mu_{Z_m^l},\sigma_{Z_m^l}) = \operatorname{Enc}_{Z}^l(Z_m^{l+1}, V_m^l),
    \label{prior}
\end{split}
\end{align}
where $\mu_{Z_m^l}$ and $\sigma_{Z_m^l}^2$ are mean and variance learned by $\operatorname{Enc}^l_Z(\cdot)$ that will be discussed in the following section. For the latent observation model, we also impose a factorized Gaussian.
Note that instead of learning its mean, we learn the observation $R^l_{n}$ directly together with its variance $\sigma^2_{R^l_n}$, which guarantees valid Gaussian conditioning during inference~\cite{volpp2020bayesian}:
\begin{align}
\begin{split}
    &p(R^l_{n} | Z^l_m, A_{m,n}) = \mathcal{N}(R^l_n | A_{m,n} Z^l_m, \operatorname{diag}(\sigma^2_{R^l_n})), \\
    &(R_n^l, \sigma_{R_n^l}) = \operatorname{Enc}_R^l(H_n^l),
\end{split}
\label{observation}
\end{align}
where $R_{n}$ and $\sigma^2_{R_n}$ are parameterized by $\operatorname{Enc}^l_R(\cdot)$.
% p(z_i) = \mathcal{N}(z_i | \mu_{z_i}, \operatorname{diag}(\sigma^2_{z_i}))
The Gaussian assumption avoids an intractable computation of the marginal likelihood of the posterior's denominator. In fact, we can calculate it easily by Gaussian conditioning in a closed-form solution. (see proof in Appendix~\ref{sec:deriGBA}):
\begin{align}
    &\bar{\sigma}_{Z_m^l}^{2}=\left[\left(\sigma_{Z_m^l}\right)^{-2}+ \sum\nolimits_{n\in \mathcal{N}^c_1(m)}\left( \sigma_{R^l_n} / A_{m,n}\right)^{-2}\right]^{-1} \label{std},\\
    &\bar{\mu}_{Z_m^l}=\bar{\sigma}_{Z_m^l}^{2} \left( \mu_{Z_m^l} / \sigma^2_{Z_m^l} + \sum\nolimits_{n\in \mathcal{N}^c_1(m)} A_{m,n} R_{n} / \sigma^2_{R^l_n} \right),
    \label{mean}
\end{align}
where $\bar{\sigma}_{Z_m^l}^{2}$ and $\bar{\mu}_{Z_m^l}$ are updated parameters and the operations are conducted in an element-wise manner. With factorization, the conditioning is efficient to compute, avoiding costly matrix inversion. In addition, all the calculations are differentiable so that GBA can be optimized in an end-to-end way by stochastic gradient descent. 

There are significant implications behind the aggregation. 
First, it incorporates the graph structure into the model by applying a linear transformation through the adjacency matrix, which is equivalent to GCNs when neglecting uncertainty terms. This suggests that GBA has similar learning abilities to GCNs.
Second, the aggregation takes the uncertainties of nodes into consideration, which is a compelling property against previous methods.
From the equations, the contribution of a context node is determined by its learned observation $R^l_n$, the variance $\sigma_{R^l_n}$, and the distance weight $A_{m,n}$. 
Specifically, Equation~\ref{std} gives a reasonable assumption that a context node located at a greater distance would provide less confident information. Additionally, Equation~\ref{mean} implicates a node's contribution diminishes when its associated variance $\sigma_{R^l_n}$ is large, signifying higher ambiguity. This theoretically guarantees the model's robustness when dealing with noisy data.
%
%In addition, the assumption of independent latent observations suggests a valid posterior regardless of the number of context nodes in Equation~\ref{posterior}, which vindicates GBA's inherent inductive ability. Thus, no sampling methods~\cite{wu2021inductive,hamilton2017inductive} are required to enhance it.

\subsection{Generative Process}
The target latent variable $Z_m^l$ depends on its representation $V_m^l$ and those of the context nodes $H^l$. 
The longer-range temporal dependencies are transited by conditioning $Z_m^l$ on $Z_m^{l+1}$, forming a vertical time hierarchy. In practice, given $V_m^l$ and a sample from $p(Z_m^{l+1})$, the network $\operatorname{Enc}^l_Z(Z_m^{l+1}, V_m^l)$ first learns a prior $p(Z_m^{l})$ over the target node in Equation~\ref{prior}.
Then, the deterministic representations of context nodes are adopted to learn their latent observations by $\operatorname{Enc}^l_R(H_n^l)$ in Equation~\ref{observation}.
Next, parameters of $p(Z_m^{l})$ are updated according to Equation~\ref{std} and \ref{mean}. After the bottom layer $l=1$, a likelihood model concatenates samples $Z_m=(Z_m^1, ...Z_m^L)$ from all layers and the target node's exogenous covariates $X_m$ to predict its extrapolations $Y_m$. Formally, the generative process of STGNP is summarized as:
\begin{align}
\begin{split}
    p(Y_m, Z_m | X_m, \mathcal{C}, A) = p(Y_m | X_m, Z_m) \prod_{l=1}^{L}p(Z_m^l | Z_m^{l+1}, V_m^l, H^l, A),
\label{generative}
\end{split}
\end{align}
where the first term is a likelihood; the second term denotes a conditional prior aggregated through the GBA. Note that at the top layer $L$, $Z_m^{L+1} = \mathbf{0}$ and the likelihood is assumed to be a factorized Gaussian distribution. 
%As there is no dependency among $z^l_{t}$ in the same layer, the generative process can also be computed in parallel.

\subsection{Inference and Optimization}
Typically, closed-form solutions for non-linear transitions and likelihood do not exist; thus we train the model through variational approximation. The approximated posterior $q(Z_m|\mathcal{C}\cup \mathcal{D}, A)$ has the same structure as the conditional prior but takes target node data $Y_m$ as inputs. Then the deterministic and stochastic modules can be optimized together by the evidence lower-bound (ELBO):
\begin{align}
\begin{split}
    \log p(Y_m|&X_m, \mathcal{C}, A) \geq \mathbb{E}_{q(Z_m)}[\log{p(Y_m| X_m, Z_m)}] \\
    &-\mathbb{KL}(q(Z_m|\mathcal{C} \cup \mathcal{D}, A)||p(Z_m|X_m, \mathcal{C}, A)).
    %&\sum_{l=1}^{L-1}\mathbb{E}_{q(z^l|\cdot)}[\mathbb{KL}()]
\end{split}
\end{align}
Given the hierarchical structure of Equation~\ref{generative}, the Kullback–Leibler divergence term $\mathbb{KL}$ can be further decomposed as:
\begin{align}
\begin{split}
    \mathbb{KL}(\cdot||\cdot) = \sum_{l=1}^L \mathbb{E}_{q(Z_m^{l+1})}
    [\mathbb{KL}(q(Z_m^{l} &| Z_m^{l+1}, {V_m^\prime}^l, H^l, A)\\ 
    &||p(Z_m^l | Z_m^{l+1}, V_m^l, H^l, A)],
    %&\sum_{l=1}^{L-1}\mathbb{E}_{q(z^l|\cdot)}[\mathbb{KL}()]
\end{split}
\label{eq:deelbo}
\end{align}
where unlike using the learned token, ${V_m^\prime}^0$ is the feature embeddings of $Y_m$.
Following~\cite{garnelo2018neural}, we use the same variational module to approximate the conditional prior so that $p(\cdot)=q(\cdot)$ in Equation~\ref{eq:deelbo}
During optimization, ELBO can be minimized using stochastic gradient descent with the reparameterization trick~\cite{kingma2013auto}.

\section{Experiments}
% \subsection{Research Questions}
To evaluate the performance and properties of STGNP, we conducted experiments to answer the following questions:
\begin{itemize}[leftmargin=*]
\item \textbf{Q1}: How does STGNP perform on real-world datasets compared to other baselines?
\item  \textbf{Q2}: What is the quality of the uncertainty estimates?
\item  \textbf{Q3}: What is the effect of each component in our model, e.g., the learnable token, CSGCN, DCConv and GBA. 
\item  \textbf{Q4}: Is STGNP resistant to different missing ratios in datasets? 
\item  \textbf{Q5}: Is it sensitive to hyperparameters and prone to overfitting?
\item  \textbf{Q6}: Does STGNP show robust generalizability when the domain of sensors changes?
\end{itemize}
\subsection{Experimental Setup}
\subsubsection{Dataset Descriptions.}
We carry out experiments on three real-world spatio-temporal datasets.
\begin{itemize}[leftmargin=*]
\item\textbf{Beijing}~\cite{zheng2015forecasting}: Beijing contains air quality indexes (AQI) from 36 stations and district-level meteorological attributes. Following~\cite{cheng2018neural,hanfine}, we aim to extrapolate the AQI of PM2.5, PM10, and NO2, using meteorological attributes such as temperature, humidity, pressure, wind speed, direction, and weather as covariates.
\item\textbf{London}\footnote{\url{https://www.biendata.xyz/competition/kdd_2018/}}: We adopt London to evaluate the performance on different domains to answer \textbf{Q6}. It collects signals from 24 stations. 
Note that some baselines use non-publicly available covariates like POIs in the above two datasets. To ensure a fair comparison, we do not utilize them for all models in this work.
%Note that for the above two datasets, the static external features like POIs adopted in previous works are not publicly available, so we do not use them.
\item\textbf{UrbanWater}~\cite{liu2016urban,liang2018geoman,liu2020predicting}: The urban water quality data comes from 15 monitoring stations in Shenzhen, which collects 3 water measures of water quality: residual chlorine (RC), turbidity (TU), and power of hydrogen (pH). Following~\cite{liu2020predicting}, we extrapolate RC given exogenous covariates TU and pH.

\end{itemize}
\begin{table*}
\centering
\caption{Performances of STGNP and the baselines on two datasets. We denote the metric variance as $\Delta_{var} = 0.1\times var$.}
  \scalebox{1.03}{
  \centering
  \begin{threeparttable}
  \begin{tabular}[width=1.\linewidth]{lcccccccccccc}
    \toprule
      \multirow{2}*{Model} & \multicolumn{3}{c}{Beijing-PM2.5} & \multicolumn{3}{c}{Beijing-PM10} & \multicolumn{3}{c}{Beijing-NO2} & \multicolumn{3}{c}{Water-RC}\\
      \cmidrule(r){2-4}  \cmidrule(r){5-7} \cmidrule(r){8-10} \cmidrule(r){11-13}
      & MAE & RMSE & MAPE & MAE & RMSE & MAPE & MAE & RMSE & MAPE & MAE & RMSE & MAPE \\
      \midrule
      KNN  & 28.08\tpm{0} & 39.87\tpm{0} & 0.62 & 61.11\tpm{0} & 99.20\tpm{0} & 0.61 & 24.18\tpm{0} & 31.36\tpm{0} & 0.95 & 0.21\tpm{0} & 0.25\tpm{0} & 0.48 \\
      IDW  & 39.11\tpm{0} & 48.90\tpm{0} & 0.73 & 72.72\tpm{0} & 116.28\tpm{0} & 0.69 & 24.15\tpm{0} & 29.76\tpm{0} & 0.96 & 0.16\tpm{0} & 0.21\tpm{0} & 0.31 \\
      RF   & 24.20\tpm{0} & 35.31\tpm{0} & 0.54 & 48.35\tpm{0} & 80.75\tpm{0} & 0.51 & 23.49\tpm{0} & 30.33\tpm{0} & 0.92 & 0.17\tpm{0} & 0.22\tpm{0} & 0.35\\
      %XGB  & 31.29\tpm{0} & 40.97\tpm{0} & 0.44 & 62.05\tpm{0} & 97.06\tpm{0} & 0.45 & 25.41\tpm{0} & 32.58\tpm{0} & 0.86 & 0.19\tpm{0} & 0.23\tpm{0} & 0.39\\
      ANCL & 19.72\tpm{5} & 30.87\tpm{8} & 0.44 & 32.43\tpm{4} & 53.15\tpm{6} & 0.31 & 20.95\tpm{3} & 26.50\tpm{6} & 0.79 & \_ & \_  & \_\\
      ADAIN& 16.81\tpm{16} & 27.00\tpm{26} & 0.32 & 31.25\tpm{13} &55.08\tpm{34} & 0.28 & 16.86\tpm{4} & 22.87\tpm{11}  & 0.54 & 0.15\tpm{0} & 0.19\tpm{0} & 0.27\\
      MCAM & 16.40\tpm{10} & 26.47\tpm{11} & 0.34 & 32.17\tpm{32} & 56.41\tpm{55} & 0.30 & 17.89\tpm{1} & 23.89\tpm{1}  & 0.63 & \_ & \_  & \_ \\
      SGNP & 17.82\tpm{1} & 28.51\tpm{9} & 0.37 & 33.76\tpm{2} & 63.96\tpm{9} & 0.29 & 17.26\tpm{1} & 22.74\tpm{0}  & 0.59 & 0.14\tpm{0} & 0.17\tpm{0}  & 0.26 \\
      SGANP & 17.06\tpm{2} & 26.74\tpm{3} & 0.33 & 31.52\tpm{3} & 55.96\tpm{6} & 0.28 & 17.42\tpm{2} & 23.01\tpm{0}  & 0.62 & 0.15\tpm{0} & 0.18\tpm{0}  & 0.26 \\
      STGNP & \textbf{14.75\tpm{3}} & \textbf{25.20\tpm{4}} & \textbf{0.28} & \textbf{27.82\tpm{3}} &\textbf{49.20\tpm{7}} & \textbf{0.26} & \textbf{15.37\tpm{1}} & \textbf{21.98\tpm{3}}  & \textbf{0.45} & \textbf{0.11\tpm{0}} & \textbf{0.16\tpm{0}}  & \textbf{0.23}\\
  \bottomrule
\end{tabular}
\end{threeparttable}
}
\label{table_main}
\end{table*}

\subsubsection{Baselines.}
We consider eight baselines which can be categorized into three classes.
\begin{itemize}[leftmargin=*]
    \item \textbf{Statistical models:} We utilize KNN, IDW~\cite{lu2008adaptive}, RF~\cite{fawagreh2014random}, and ANCL~\cite{patel2022accurate}. ANCL is a GPs-based framework that designs specialized kernels for different data attributes.
    \item \textbf{Neural Network methods:} We take {ADAIN}~\cite{cheng2018neural} and {MCAM}~\cite{hanfine} as NNs baselines. ADAIN uses MLP and RNN layers for static and dynamic data, followed by an attention mechanism to aggregate features, whereas MCAM introduces multi-channel attention blocks for static and dynamic correlations. 
    \item \textbf{Neural Processes approaches:} We modify SNP~\cite{singh2019sequential} and name the variant as SGNP. As SNP cannot deal with graph data, we add a cross-set graph network at each recurrent step before the aggregation. SGANP is an advanced version of SGNP modified from \cite{qin2019recurrent}, which utilizes attention mechanisms as the aggregator.
\end{itemize}

Our model, STGNP, is based on a probabilistic framework and is a general method that is not tailored to specific tasks. It can also be applied in situations where exogenous covariates are not available by simply removing the corresponding causal convolution blocks.
In contrast, ANCL relies on periodic and categorical kernels and MCAM utilizes horizontal and vertical wind speeds that are specific to the air quality task. 
SGNP and SGANP have a similar NP framework to ours, but we abandon the recurrent structure and propose GBA to aggregate nodes under a Bayesian framework.

\begin{figure}[b]
  \centering
  \includegraphics[width=1.0\linewidth]{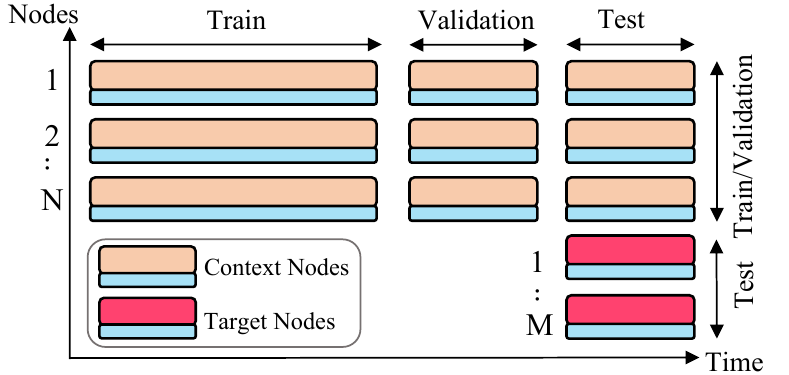}
  \caption{Evaluation strategy and dataset division. During testing, we use context nodes to extrapolate target nodes.}
    \label{figure_dataset_division}
\end{figure}
\subsubsection{Evaluation Strategy and Hyperparameters.}

In Figure~\ref{figure_dataset_division}, we illustrate the evaluation strategy for experiments. As we lack fine-grained data for the areas, we manually set aside 30\% of existing stations in the dataset as target nodes for reporting performance. The remaining 70\% is used for training purposes, ensuring a ratio of $3N=7M$. This ensures that target nodes will not involve in the training phase.
% We sequentially split the data into three segments: a training set consisting of the first 80\% of the data, a validation set consisting of the next 10\%, and a test set consisting of the final 10\%.
The data is sequentially divided into three segments: an 80\% training set, a 10\% validation set, and a 10\% test set.
During training and validation, we use $N-3$ nodes to extrapolate the remaining $3$ randomly selected nodes.
We report extrapolation metrics of MAE, MSE, and MAPE of the target stations. Since our model, SGNP, and SGANP contain stochastic modules, we adopt the mean of distributions directly, instead of sampling, to report the results.
The time window $T$ is $24$ and the adjacency matrix is constructed based on the locations of stations, normalized by a thresholded Gaussian kernel~\cite{shuman2013emerging}. 
It is worth noting that, in the case of NNs, data processing is necessary to handle missing values in the dataset. For this reason, we employ linear interpolation to fill in the missing values for ADAIN and MCAM, while for the other baselines, missing values are left as zero.
Hyperparameters are the same on all datasets for STGNP. The deterministic learning stage has $3$ layers with a kernel size $k=3$ and channel numbers $[16, 32, 64]$. The stochastic stage is a 3-layer $1\times 1$ convolutions module and the channels of latent variables are $[16, 32, 64]$. The likelihood function is a 3-layer $1\times 1$ convolution with $128$ channels in each layer. STGNP, ADAIN, MCAM, SGNP, and SGANP are implemented with PyTorch and trained on an NVIDIA A100 GPU. We repeat each experiment $5$ times and report the average and variance of the metrics. Our implementations\footnote{\url{https://github.com/hjf1997/STGNP}} are publicly available.

\subsection{Overall Performance}
%How is the performances of STGNP on real-world datasets among other baselines
To answer \textbf{Q1}, we report the overall results of baselines over Beijing and Water datasets, as shown in Table~\ref{table_main}. Note that ANCL and MCAM require meteorological features, so they cannot be applied to the Water dataset.
From the table, we see that STGNP consistently outperforms other baseline models with a notable margin, with the lowest errors on all datasets. Moreover, we have the following observations. First, the GPs model ANCL outperforms the other statistical models, indicating that GPs have the essential ability to capture complex dependencies with dedicated kernels. 
Second, NNs models surpass the above methods on all datasets because of their strong learning capability. Third, we find that SGNP and SGANP cannot consistently outperform NNs, possibly due to the transition collapse of the recurrent structure which may cause them to ignore the input contexts. Comparing these two models, although the attention aggregator performs better than the mean aggregator on tasks like computer vision, this is not always the case for graph data. This is because the data is constrained by a graph, which should be considered explicitly. Our STGNP has the best results due to the use of causal convolutions to alleviate transition collapse and the GBA to take into fact node uncertainties. 

Another interesting discovery is that our model, SGNP, SGANP, and ANCL exhibit lower metric variances compared to other NN models. Evidently, for PM10 concentrations with large extrapolation errors, the metric variances for these models are small whereas for ADAIN and MCAM, their MAE and RMSE reach high values of $1.3$, $3.4$, and $3.2$, $5.5$, separately. The primary reason for the models with low variances is their ability to model data in a probabilistic manner and their awareness of uncertainty. This characteristic enables the models to be robust to parameter initialization and reduces the chance of getting trapped in local minima.
%The chief reason is that the models with low variances can model data in a probabilistic fashion and are uncertainty aware. This makes the models robust to parameter initialization and less likely to be trapped in local minima.

\begin{figure}[!b]
  \centering
  \includegraphics[width=1.03\linewidth]{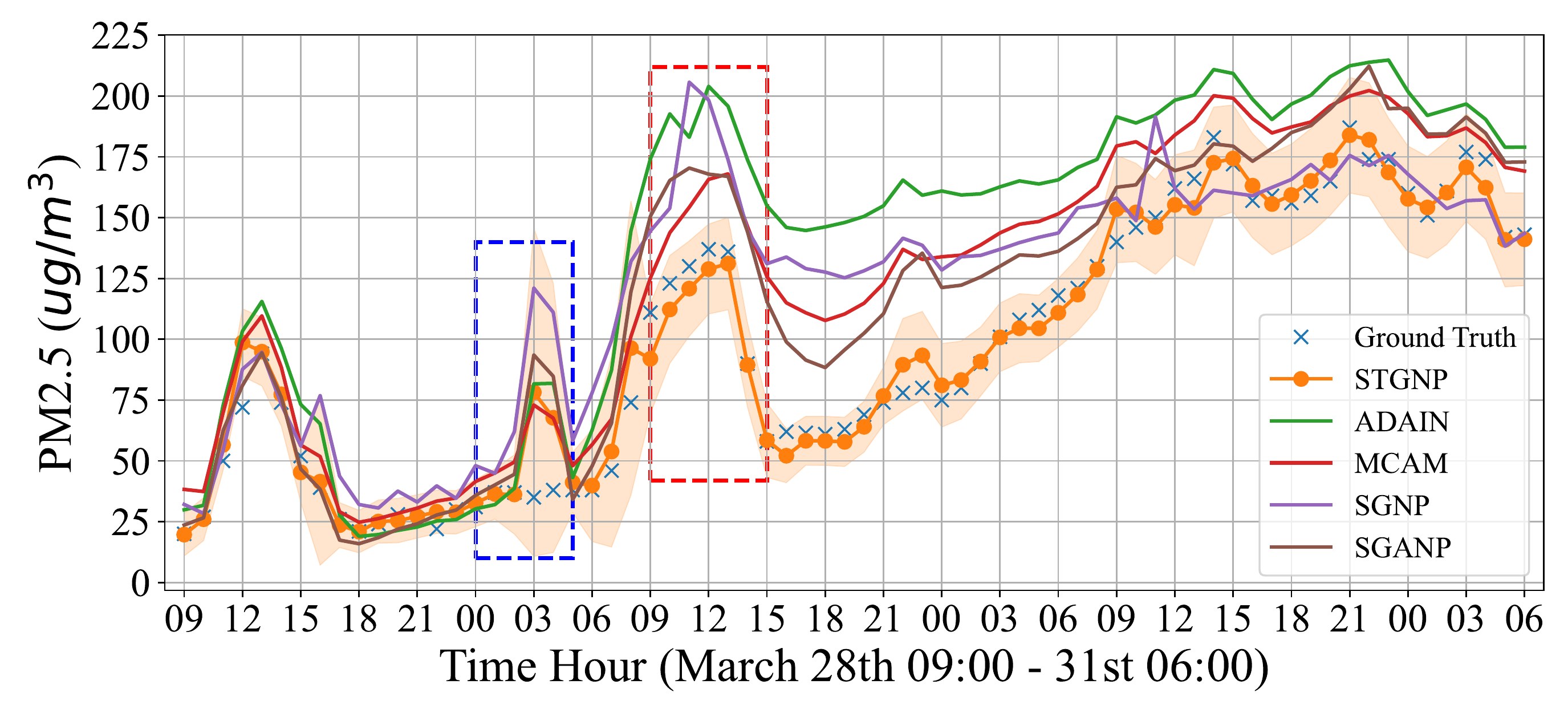}
  \caption{PM2.5 extrapolation results of the Beijing dataset during March 28 -- 31, 2015.}
    \label{figure_results_visualization}
\end{figure}

We also visualize extrapolation results in Figure~\ref{figure_results_visualization} which illustrates the PM2.5 extrapolations of the five best models on the Beijing dataset, together with the ground truth. We observe that our model produces the most accurate results toward the ground truth data. This is particularly evident during the sudden change in the data after 9 AM on March 29, as highlighted in the red box. This demonstrates the effectiveness of our model in capturing both spatial information from surrounding sensors and temporal dependencies.

\subsection{Uncertainty Estimates Analysis}
One of the key benefits of our STGNP model is its capability to provide high-quality uncertainty estimates. To answer \textbf{Q2}, we first evaluate the qualitative results to show that they provide valuable information. In Figure~\ref{figure_results_visualization}, our model consistently has more accurate extrapolations compared with baselines. However, at 3--6 AM, March 28 shown in the blue box, all methods fail to generate precise data. Significantly, our model renders higher uncertainties (orange shadow), indicating possible inaccurate extrapolations. This ability to accurately estimate uncertainty can be extremely useful in practical decision-making. For instance, if the uncertainty for a location is consistently high, researchers could use this information to prioritize the deployment of sensors in order to achieve more accurate data analysis with minimal expenditure.

Next, to evaluate the quality of our estimates quantitatively, we follow the approach used in~\cite{chai2018bike} and examine the number of extrapolations that fall in the uncertainty intervals. Assume a Gaussian likelihood with variance $\sigma^2$ to suggest uncertainty, the intervals $1\sigma, 2\sigma, 3\sigma$ centered at the predicted data cover $\approx$ $68.3\%, 95.5\%, 99.7\%$ of the probability density. We posit that a better model should have a higher proportion of points falling in the intervals. Table~\ref{table_uncertainty} reports the statistical results of the proportions for ANCL, SGNP, SGANP, and STGNP. It shows that all NPs methods have superior performance compared to ANCL and that our STGNP outperforms SGNP and SGANP in most metrics. This is likely due to the fact that our model also considers uncertainties in context nodes through GBA.
%that our STGNP has overall best estimation abilities compared to SGNP and ANCL models.

\begin{table}[!h]
\caption{Proportions (\%) of data falling in the intervals of $1\sigma-3\sigma$, where $\sigma$ is the standard deviation of the likelihood.}
  \scalebox{0.90}{
  \centering
  \begin{threeparttable}
  \begin{tabular}[width=0.8\linewidth]{lcccc}
    \toprule
      \multirow{2}*{Model} & Beijing-PM2.5 & Beijing-PM10 & Beijing-NO2 & Water-RC\\
      \cmidrule(r){2-2} \cmidrule(r){3-3} \cmidrule(r){4-4} \cmidrule(r){5-5}
      & $1\sigma$ / $2\sigma$ / $3\sigma$ & $1\sigma$ / $2\sigma$ / $3\sigma$ & $1\sigma$ / $2\sigma$ / $3\sigma$ & $1\sigma$ / $2\sigma$ / $3\sigma$\\
      \midrule
      ANCL & 44 / 67 / 86 & 16 / 30 / 49 & 35 / 70 / 88 & \_ \\
      SGNP & 71 / 92 / 96 & 72 / 91 / 97 & 71 / \textbf{95} / \textbf{98} & 62 / 90 / 95\\
      SGANP & 73 / 92 / \textbf{97} & 78 / 93 / 97 & 69 / 91 / 96 & 61 / 92 / 97\\
      STGNP & \textbf{77} / \textbf{92} / \textbf{98} & \textbf{85} / \textbf{96} / \textbf{98} & \textbf{76} / 94 / \textbf{98} & \textbf{66} / \textbf{94} / \textbf{99}\\
  \bottomrule
\end{tabular}
\end{threeparttable}
}
\label{table_uncertainty}
\end{table}

\begin{figure*}[!t]
\setlength{\abovecaptionskip}{-0.01mm}
\centering
\subfigcapskip=-2pt
\subfigure[Missing ratios of data]{
\begin{minipage}[t]{0.25\linewidth}
\centering
\includegraphics[width=0.95\linewidth]{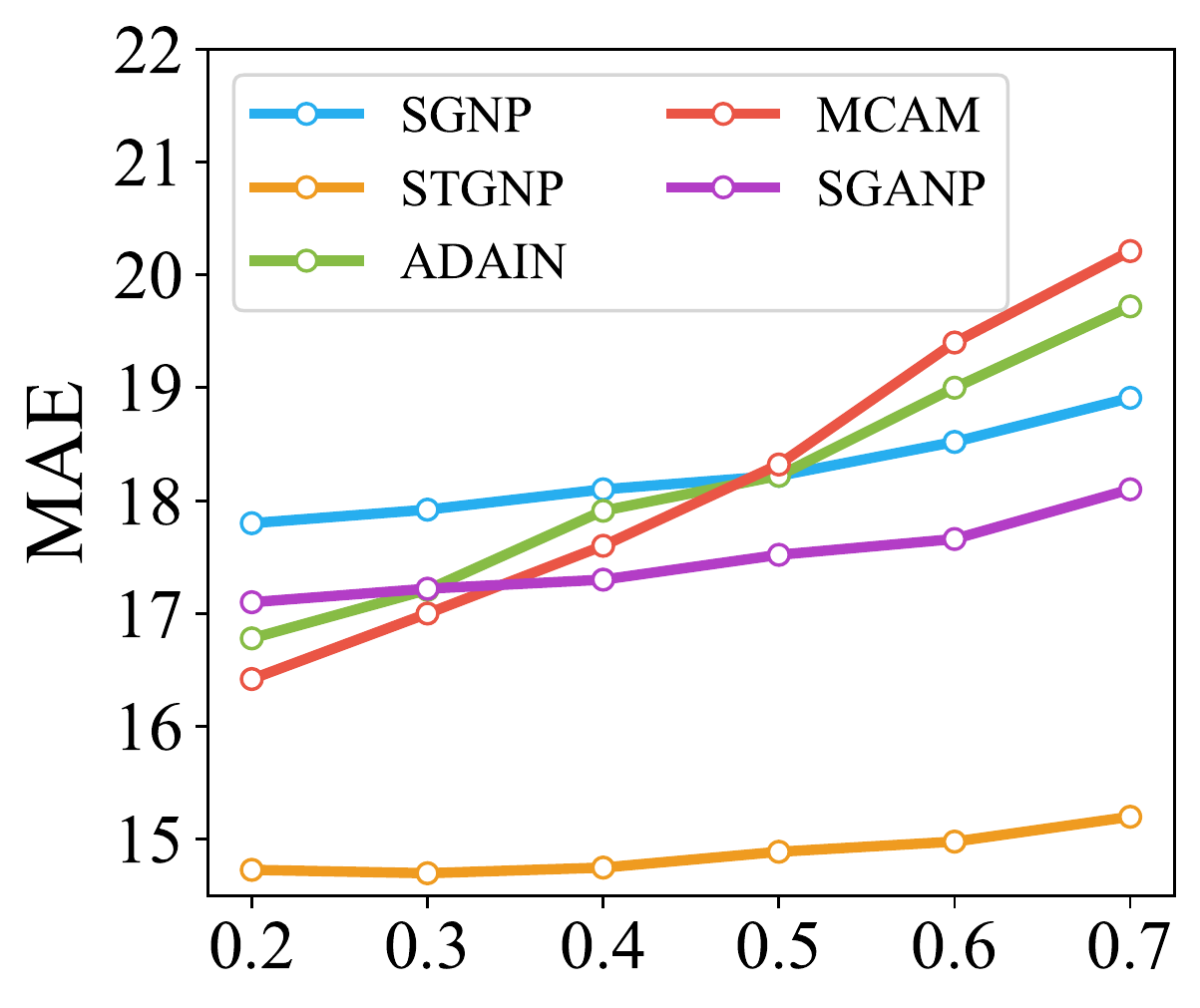}
%\caption{fig1}
\end{minipage}
}%\hskip -4mm
\subfigcapskip=-2pt
\subfigure[Channels of Stages w.r.t. $u$]{
\begin{minipage}[t]{0.25\linewidth}
\centering
\includegraphics[width=1.02\linewidth]{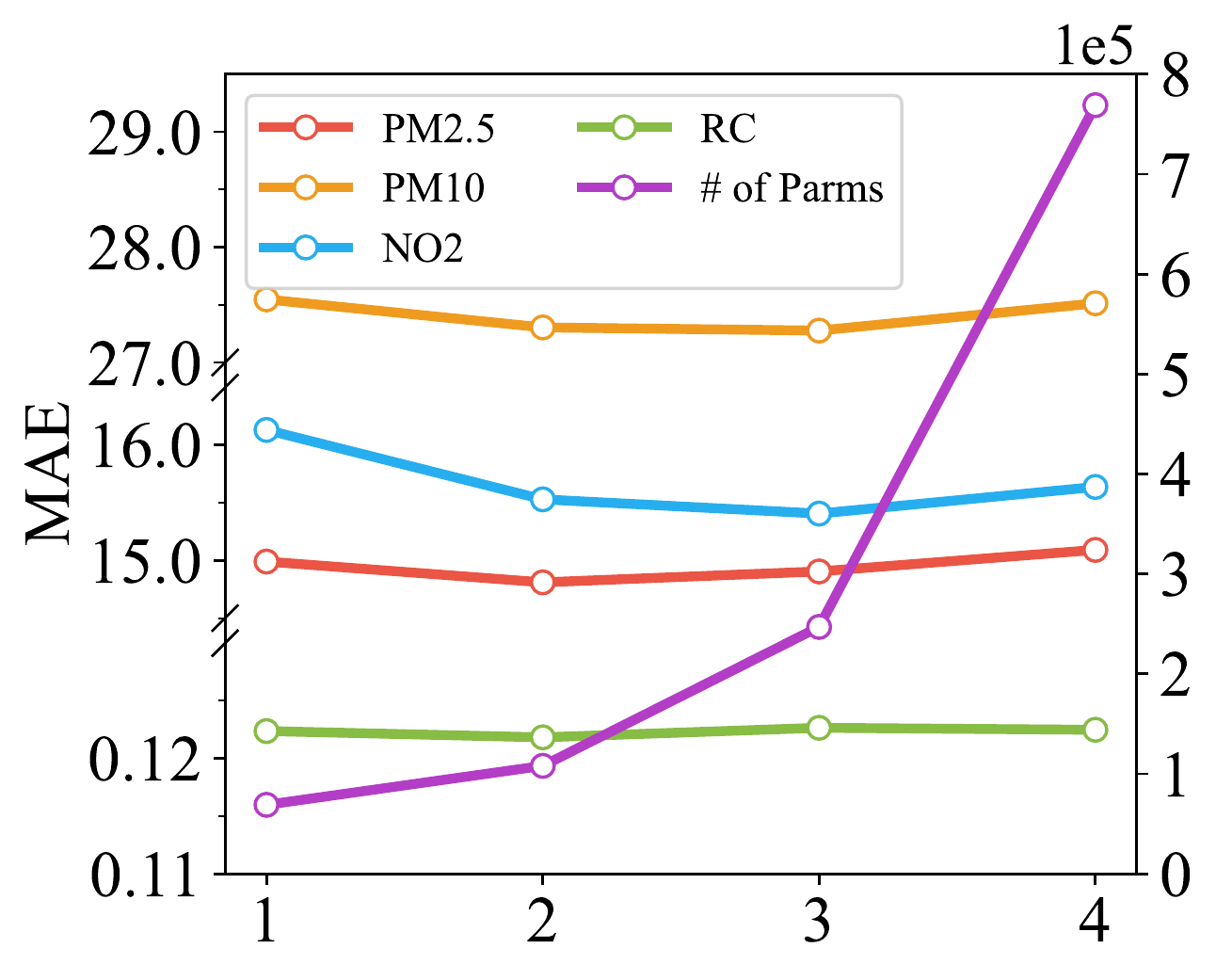}
%\caption{fig1}
\end{minipage}
}%\hskip -4mm
\subfigcapskip=-2pt
\subfigure[Channels of likelihood module]{
\begin{minipage}[t]{0.25\linewidth}
\centering
\includegraphics[width=1.0\linewidth]{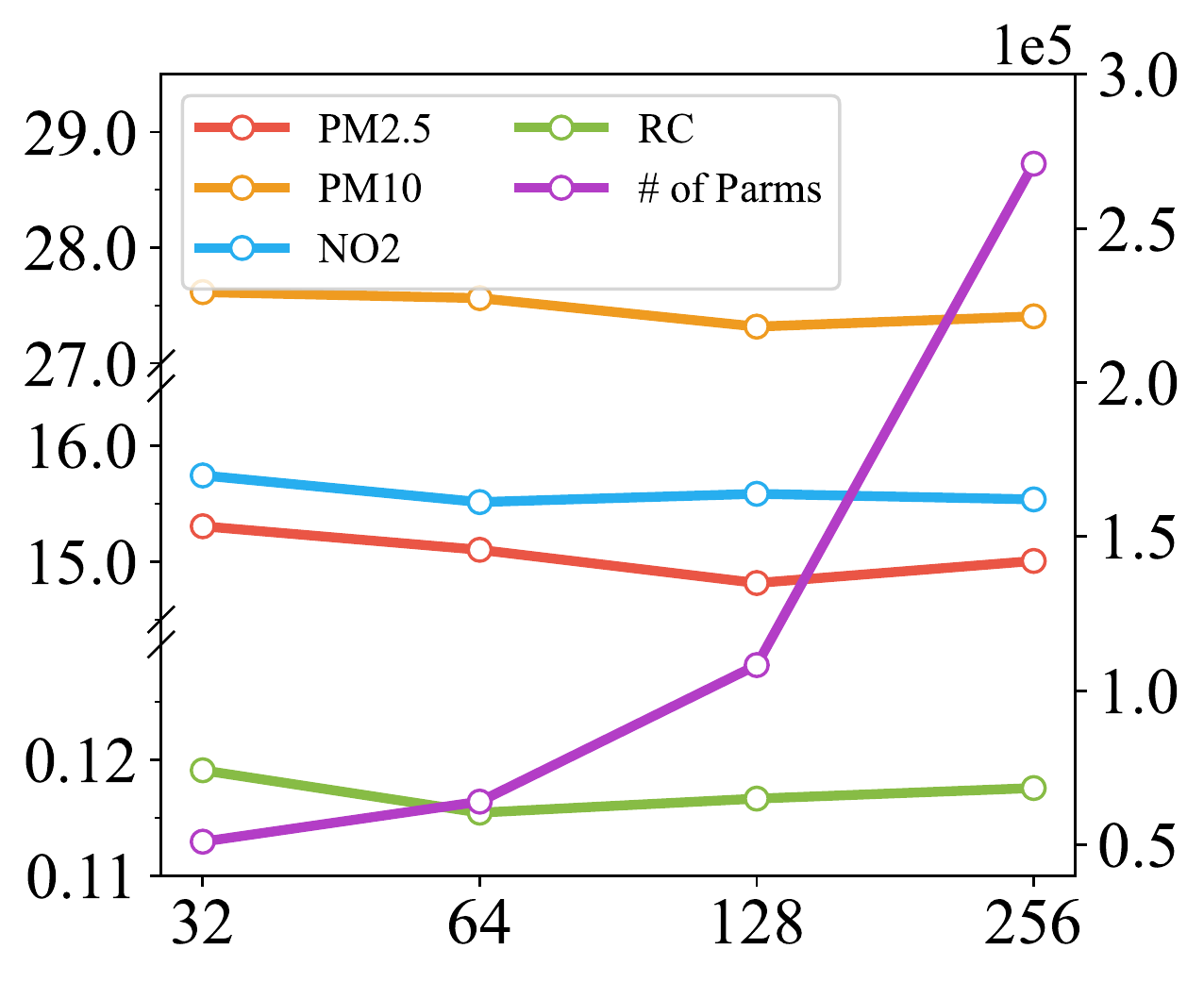}
%\caption{fig1}
\end{minipage}
}%\hskip -4mm
\subfigcapskip=-2pt
\subfigure[Number of STGNP Layers]{
\begin{minipage}[t]{0.25\linewidth}
\centering
\includegraphics[width=1.02\linewidth]{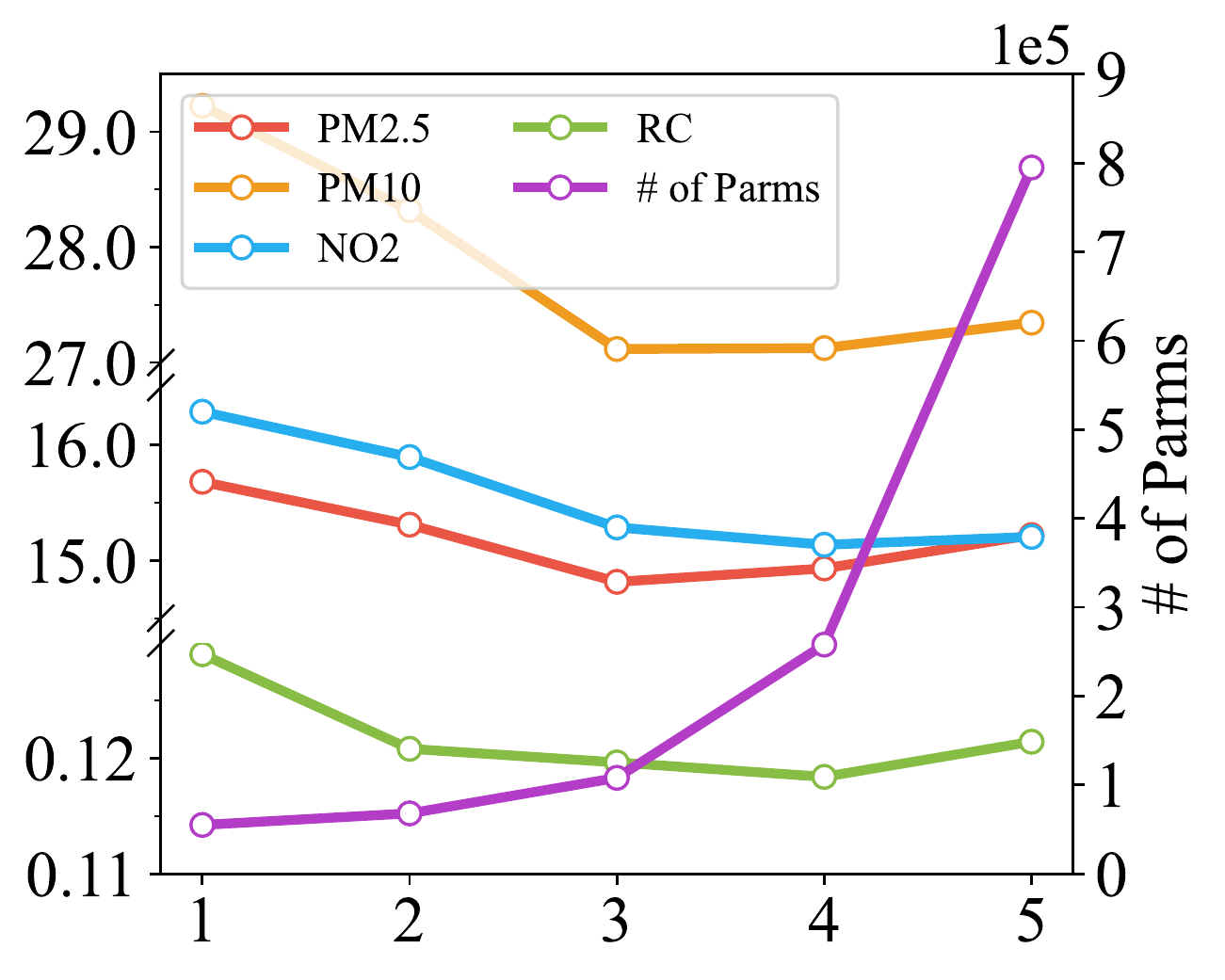}
\end{minipage}%
}%
\centering
\caption{Data missing ratio and hyperparameter study. In (b), channel numbers are $[8u, 16u, 32u]$ for causal convolutions and latent variables with $u$ ranging from $1$ to $4$.}
\label{figure_params}
\end{figure*}

\subsection{Ablation Study}
\label{sec:ablation}
\begin{figure}[!b]
  \centering
  \includegraphics[width=0.98\linewidth]{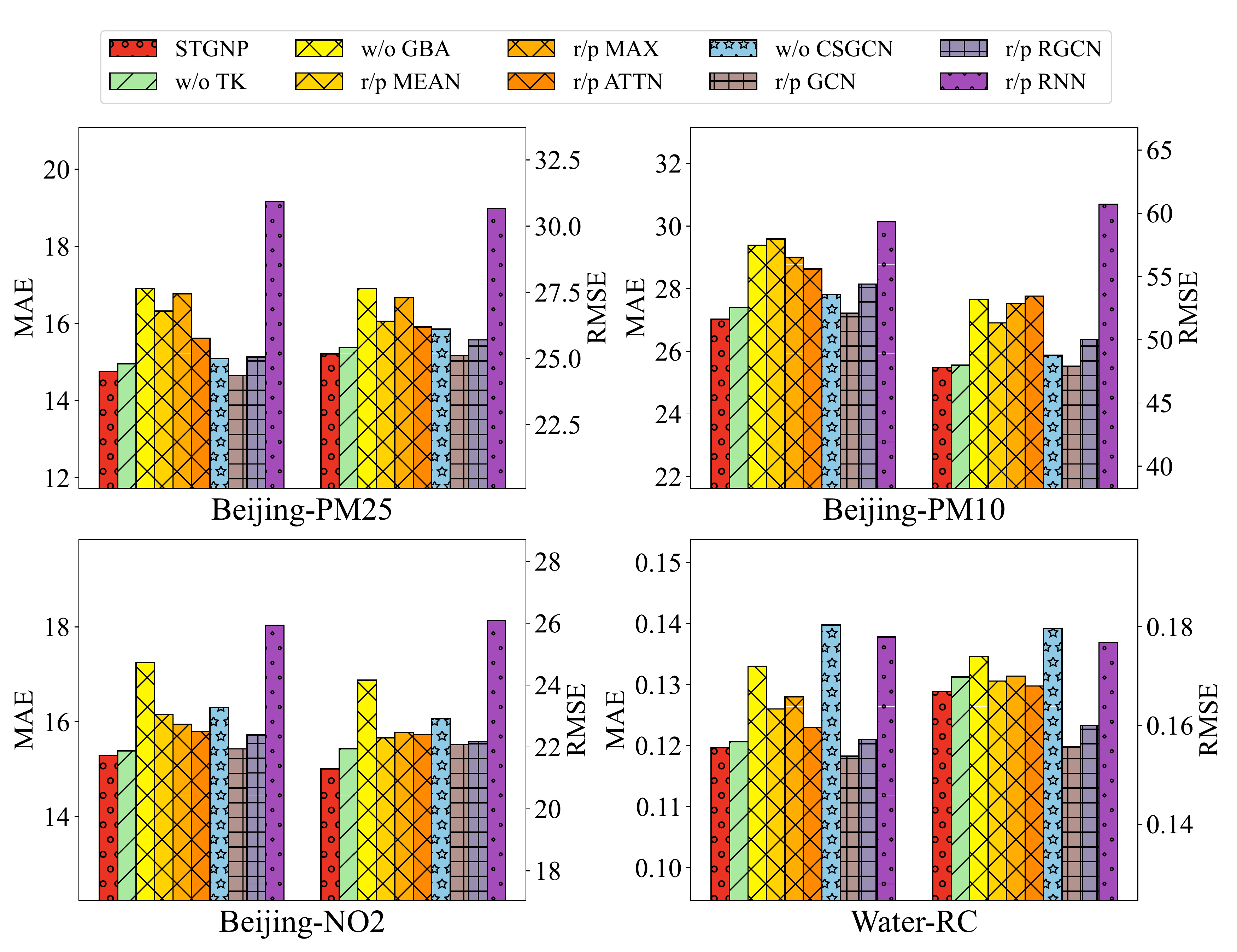}
  \caption{Effectiveness of different modules in STGNP.}
    \label{figure_ablation}
\end{figure}

To assess the contribution of each component to the overall performance of our model and answer \textbf{Q3}, we conduct ablation studies and present the results in Figure~\ref{figure_ablation}. In each study, we only modify the corresponding part while leaving other settings unchanged.

\textbf{Effect of learnable token}: We remove the learned token and use linear interpolation to preprocess target nodes (w/o TK). The results unveil that the token has a positive impact on the model's performance. We believe this is because using simple interpolation to represent all signals of the target nodes leads to significant errors. 
Mingled with context nodes' signals, this hobbles the model's learning ability, and even the uncertainty estimates struggle to rectify them. In contrast, the learned token identifies a suitable position in the feature space that accurately indicates the target nodes, thereby avoiding this problem.

\textbf{Effect of CSGCN}: To evaluate the effectiveness of cross-set graph convolutional network, we compare it to three variants: (a) w/o CSGCN: removing CSGCN in our model; (b) r/p GCN: replacing CSGCN with a standard GCN. (c) r/p RGCN: replacing it with an advance relational GCN~\cite{schlichtkrull2018modeling}. Our results show that removing the spatial learning module CSGCN leads to a degradation in performance, underscoring the importance of capturing spatial dependencies. Additionally, we find our CSGCN achieves performance on par with the standard GCN, which learns dependencies among all nodes in two sets, and even slightly better than the RGCN, which explicitly characterizes categorical relations among nodes. This observation suggests that modeling dependencies across two sets are sufficient to achieve satisfactory performance, thereby validating our insight.

\textbf{Effect of GBA}: Results from experiments where GBA is discarded (w/o GBA) or replaced with a max aggregator (r/p MAX), a mean aggregator (r/p MEAN), or an attention aggregator (r/p ATTN) highlight the critical role of GBA. Removing or replacing it leads to a significant decline in results on most datasets. This is likely because context nodes have varying levels of ambiguities. Without uncertainty estimates, the model has difficulty extracting valuable context information, which hampers the performance of the likelihood module.

\textbf{Effect of causal convolution}: We replace our temporal learning framework with an RNN structure (r/p RNN) and the results show a strong deterioration in the model's performance. The outcome indicates transition collapse, which occurs in an RNN because of a large number of transitions, can have a substantial impact on the model's capability. In contrast, our STGNP model effectively mitigates this issue by significantly reducing the number of transitions.

% \begin{figure}[!b]
% \setlength{\abovecaptionskip}{-0.01mm}
% % \setlength{\belowcaptionskip}{-12pt}
% \centering
% \subfigure{
% \begin{minipage}[t]{0.5\linewidth}
% \centering
% \includegraphics[width=1.07\linewidth]{figure/ablation_study/ablation(a).pdf}
% %\caption{fig1}
% \end{minipage}%
% }%\hskip -8pt
% \subfigure{
% \begin{minipage}[t]{0.5\linewidth}
% \centering
% \includegraphics[width=1.08\linewidth]{figure/ablation_study/ablation(b).pdf}
% % (b) Size of hidden states for ASGGRU.
% %\caption{fig2}
% \end{minipage}
% }\vskip -10pt

% \subfigure{
% \begin{minipage}[t]{0.5\linewidth}
% \centering
% \includegraphics[width=1.07\linewidth]{figure/ablation_study/ablation(c).pdf}
% % (c) Number of input sequence length $T$.
% %\caption{fig2}
% \end{minipage}
% }%\hskip -10pt
% \subfigure{
% \begin{minipage}[t]{0.5\linewidth}
% \centering
% \includegraphics[width=1.07\linewidth]{figure/ablation_study/ablation(d).pdf}
% % (c) Number of short-term frames $T_s$.
% %\caption{fig2}
% \end{minipage}
% }%
% \centering
% \caption{Effectiveness of different modules in STGNP.}
% \label{figure_ablation}
% \end{figure}

\subsection{Missing Ratio Study}

Evaluating the model's performance on data with missing values is important, as real-world sensors might lose signals due to unpredictable factors.
To answer \textbf{Q4}, we train models using manually corrupted data by randomly replacing a ratio of data with zero. Following the same procedure, we preprocess them with linear interpolation for NNs, while leaving the missing values as zero for NPs models. The results, which illustrate the MAE performance of various baselines for ratios ranging from 0.2 to 0.7, are shown in Figure~\ref{figure_params}(a). We observe significant performance degradation in ADAIN and MCAM, which can be similarly attributed to large errors caused by interpolation.  In contrast, STGNP and SGANP demonstrate better performance, with STGNP performing the best results.
The key factor behind their success lies in the ability of their likelihood modules to capture uncertainties associated with the target nodes. This capability effectively mitigates the impact of missing data.
Compellingly, the GBA module in our STGNP is able to further enhance its robustness by accurately modeling uncertainties in the signals of individual sensors.

\subsection{Hyperparameter Study}
To answer \textbf{Q5}, we study the performance of STGNP under various hyperparameter settings. We first explore the impact of the number of channels in two stages. Following the default setting of the channel numbers $[8u, 16u, 32u]$ where $u=2$, we vary $u$ from $1$ to $4$. From Figure~\ref{figure_params}(b), we observe that even with the drastically increasing number of learnable parameters (violet curves), the MAE metrics (green, blue, red, and yellow curves) keep stable. 
Then, we examine the effect of changing the number of channels of the likelihood module ranging from $32$ to $256$ and also observe stable performances, as shown in Figure~\ref{figure_params}(c). 
Finally, we investigate the effect of the number of layers ranging from $1$ to $5$. From Figure~\ref{figure_params}(d), the performances initially improve as it increases, but start to oscillate since $3$ layers. 
This can be explained by the perceptive field of the model, where 1 and 2 layers correspond to fields of $3$ and $7$, respectively. These limited perceptive fields constrain the model's performance.
However, once the perceptive field becomes sufficiently large (i.e., at 3 layers and above), the model captures long enough temporal dependencies, leading to improved and more stable performance. These findings suggest STGNP is insensitive to hyperparameters, as long as the perceptive field is adequately large.

\subsection{Cross-Domain Evaluation}
To assess the generalization ability of models and answer \textbf{Q6}, we train them on the Beijing dataset and evaluate results using the London dataset. 
Table~\ref{table_transfer} reports the performances of PM2.5, where the first and second columns of MAE mean the performance of training models using London data directly and training on Beijing while evaluating on London. We confirm that our STGNP has the best results in both training approaches. In addition, when measuring the performance discrepancy, we also discover that STGNP has the smallest performance degeneration, especially compared to NNs models. This is likely due to STGNP's strong spatio-temporal learning capability as well as its non-parametric design inherited from GPs. This allows the model to learn high-level spatio-temporal principles in the feature space and to instantiate these dependencies during testing by constructing a stochastic process from the sets.
\begin{table}[!h]
\caption{Performances of cross-domain evaluation.}
  \scalebox{0.94}{
  \centering
  \begin{threeparttable}
  \begin{tabular}[width=0.8\linewidth]{lccc}
    \toprule
      \multirow{2}*{Model} & \multicolumn{3}{c}{Beijing$\rightarrow$London-PM2.5} \\
      \cmidrule(r){2-4}
      & MAE & RMSE & MAPE  \\
      \midrule
      ADAIN& 4.35/3.57 ($\downarrow$0.78) & 5.79/4.69 ($\downarrow$1.10) & 0.62/0.42 ($\downarrow$0.20) \\
      MCAM & 4.38/3.78 ($\downarrow$0.60) & 5.38/4.80 ($\downarrow$0.58) & 0.68/0.49 ($\downarrow$0.19)  \\
      SGNP & 4.31/4.11 ($\downarrow$0.20) & 5.41/5.12 ($\downarrow$0.29) & 0.66/0.60 ($\downarrow$0.06)\\
      SGANP & 3.85/3.61 ($\downarrow$0.24) & 5.15/4.82 ($\downarrow$0.33) & 0.52/0.45 ($\downarrow$0.07)\\
      STGNP & \textbf{3.20/3.03 ($\downarrow$0.17)} & \textbf{4.30/4.26 ($\downarrow$0.04)} & \textbf{0.40/0.33 ($\downarrow$0.07)}\\
  \bottomrule
\end{tabular}
\end{threeparttable}
}
\label{table_transfer}
\end{table}

\section{Related Work}
\subsection{Spatio-Temporal Extrapolation}
Spatio-temporal extrapolation is a task that involves predicting the state of a surrounding environment based on known information. Early works in this area use statistical machine learning methods, such as $K$-Nearest Neighbors (KNN) and Random Forest (RF)~\cite{fawagreh2014random}, to solve this problem. KNN approaches rely on linear dependencies between data points, while RF can capture non-linear dependencies. However, these methods only consider spatial relations and struggle to model more complex and dynamic correlations.
Gaussian Processes~\cite{seeger2004gaussian} learn to construct stochastic processes, in which the spatio-temporal dependencies are captured by flexible kernels that are designed to handle different types of features~\cite{li2020stochastic}. For instance, Patel \emph{et al.}~\cite{patel2022accurate} used periodical and Hamming distance kernels for temporal and categorical features respectively. However, these kernels are tailored to specific scenarios and the cubic time complexity limits their applicability.
Some approaches view extrapolation as a tensor completion task~\cite{yu2016temporal}. Based on a low-rank matrix assumption, these methods capture the spatio-temporal patterns while being efficient to optimize. However, they are transductive, meaning that they can only infer the state of nodes involved in the training process. They are not able to generalize to new nodes that were not present in the training data.
Recently, Neural Networks (NNs) have become the dominant paradigm. 
Cheng \emph{et al.}\cite{cheng2018neural} proposed an attention model for air quality inference, where the dynamic and static data are encoded by RNNs and MLPs, and the attention mechanism integrates the features of nodes. Han \emph{et al.} \cite{hanfine} combined GCNs with a multi-channel attention module to improve performance. However, NNs tend to struggle with learning uncertainties and can overfit on datasets with low amounts of data.
Some NNs also view extrapolation as a kriging problem. For instance, Appleby \emph{et al.} \cite{appleby2020kriging} interpolate a node given its neighbors with time information as extra features, and We \emph{et al.} \cite{wu2021inductive} learn the temporal dynamics, explicitly. However, the first only captures spatial relations and the second cannot incorporate exogenous covariates. On the contrary,  our approach is able to achieve both goals.

%The difference is that kriging only depends on the graph structure to infer node signals. Instead, we focus on a more general setting of using nodes' existing input features. In fact, our model can simply be applied to kriging by replacing the target node prior with static modeling based on its graph structure.

% we focus on a general setting utilizing node's input features and the graph structure, while the kriging model can only model    

\subsection{Neural Processes Family}
Neural Processes (NPs) combine the merits of both NNs and GPs~\cite{garnelo2018neural}, which possess strong learning ability and uncertainty estimates. 
Basically, it induces latent variables over the context set, forming a conditional latent variable model. Then, a likelihood module is utilized to generate the target outputs. Le \emph{et al.}~\cite{le2018empirical} found that NPs suffer an underfitting problem because of the incapable aggregation function (e.g., mean or sum). 
Then, Kim \emph{et al.}~\cite{kim2019attentive} proposed Attentive Neural Processes (ANP) to calculate the importance within/across the context set and target set. 
Kim \emph{et al.}~\cite{kim2022neural} further proposed a stochastic attention mechanism for aggregation where the attention weights are inferred using Bayesian inference and Volpp \emph{et al.}~\cite{volpp2020bayesian} introduced a stochastic aggregator to aggregate context variables directly. However, these works only consider the spatial domain and cannot handle graph data.
Singh \emph{et al.}~\cite{singh2019sequential} first focused on sequential data and proposed Sequential Neural Processes (SNP). With a latent state transition function from a variational recurrent neural network (VRNN)~\cite{chung2015recurrent}, it constructs a sequential stochastic process for a timeline. Then, Yoon \emph{et al.}~\cite{yoon2020robustifying} introduced Recurrent Memory Reconstruction to compensate for the distribution shift in a sequence. However, these recurrent structures have the transition collapse problem, which can make it difficult to learn temporal relations over long sequences. In contrast, our method utilizes casual convolutions to alleviate the challenge.

\section{Conclusions and Future Work}
We introduce Spatio-Temporal Graph Neural Processes, the first framework for spatio-temporal extrapolation in the Neural Processes family. Our model captures temporal relations and addresses the transition collapse problem using causal convolutions while effectively learning spatial dependencies using the cross-set graph network.
The Graph Bayesian Aggregation aggregates context nodes in a way that takes into account their uncertainties and enhances the learning ability of NPs on graph data. 
Experimental results demonstrate the superiority of STGNP in terms of extrapolation accuracy, uncertainty estimates, robustness, and generalizability.
In the future, an intriguing direction would be to explore STGNP for spatio-temporal forecasting, which is another fundamental task in the area. By aggregating historical representations, the model could provide predictions about future time steps.

\begin{acks}
This research is supported by Singapore Ministry of Education Academic Research Fund Tier 2 under MOE's official grant number T2EP20221-0023. It is also supported by Guangzhou Municipal Science and Technology Project 2023A03J0011.

\end{acks}

\bibliographystyle{ACM-Reference-Format}
\balance
\bibliography{sample-base}

\clearpage
\appendix

\section{Mathematical Notation}
\label{sec:notation}
We define the major mathematical notations in the paper in Table~\ref{table_notation} for better understanding. 
\begin{table}[!h]
\centering
\caption{Major notations used in the paper.}
\label{table_notation}

  \begin{threeparttable}
  \begin{tabular}[width=0.69\linewidth]{lll}
    \toprule
      Notation & Dimension & Description \\
      \midrule
    $N$, $M$ & $\mathbb{R}^1$ & number of context, target nodes\\
    $T$ & $\mathbb{R}^1$ & time length of a sequence\\ 
    $A_{m,n}$ & $\mathbb{R}^1$ & weight between node $m$ and $n$\\
    $n$, $m$  & $\mathbb{R}^1$ & index of a context and target node\\
    $d_l$, $d_x$, $d_y$ & $\mathbb{R}^1$ & feature dimensionalities\\
    \midrule
    $\mathcal{C}$ & $\mathbb{R}^{N\times T\times (d_x+d_y)}$ & context set\\
    $\mathcal{D}$ & $\mathbb{R}^{M\times T\times (d_x+d_y)}$ & target set\\
    $H$ & $\mathbb{R}^{M\times T\times \sum_{l=1}^Ld_l)}$ & representations of context nodes\\
    \midrule
    $X_n$, $Y_n$ & $\mathbb{R}^{T\times (d_x+d_y)}$ & covariates, data of a context node\\
    $X_m$, $Y_m$ & $\mathbb{R}^{T\times (d_x+d_y)}$ &  those of a target node\\
    $H_n$ & $\mathbb{R}^{T\times \sum_{l=1}^Ld_l}$ &  representations of context node $n$\\
    $V_m$ & $\mathbb{R}^{T\times \sum_{l=1}^Ld_l}$ & representations of target node $m$\\
    $R_n$ & $\mathbb{R}^{T\times \sum_{l=1}^Ld_l}$ & latent observations of node $n$\\
    $Z_m$ & $\mathbb{R}^{T\times \sum_{l=1}^Ld_l}$ & latent variables of target node $m$\\
  \bottomrule
\end{tabular}
\end{threeparttable}
\end{table}

\section{Derivation of Graph Bayesian Aggregation}
\label{sec:deriGBA}
In this section, we give formal derivations of the proposed Graph Bayesian Aggregation. We first derive the general GBA without factorization or specific graph stricture. We assume a latent prior $Z$ over a target node (omitting subscript $m$ for brevity). The latent observation functions of all context nodes are an independent linear transformation of $Z$ following Gaussian distributions: 
\begin{align}
&p(Z)=\mathcal{N}(Z|\mu,\Lambda^{-1}),\\
&p(R_n|Z)=\mathcal{N}(R_n|A_n Z,L_n^{-1})\quad\text{  For } n \in [1, N],
\end{align}
where we use the precision matrix $\Lambda$ and $L$ for convenience; $A_n$ is a transformation matrix, representing the graph structure. The logarithmic joint probability over $Z$ and $[R_1, .., R_N]$ is:

\begin{eqnarray}    \label{eq15}
\ln{p(Z, R_1, .., R_N)}&=&\ln{p(Z)}+\ln{p(R_{1} ...R_{N}|Z)}    \\
% ~&=&\ln{p(Z)} + \sum_{n=1}^{N} \ln{p(R_{n}|Z) } \nonumber    \\
&=&-\frac{1}{2} (Z-\mu )^T\Lambda(Z-\mu)\nonumber\\
&&\quad-\frac{1}{2}\sum_{n=1}^{N}(R_{i}-A_iZ)^TL_{i}(R_i-A_iZ)+\operatorname{Const}, \nonumber
\end{eqnarray}
where the second order terms can be decomposed as:
\begin{eqnarray}    
-\frac{1}{2}\begin{bmatrix}
    Z\\
    R_{1}\\
    \vdots \\
R_{N}
\end{bmatrix} 
\underbrace{
\begin{bmatrix}
    \Lambda+\sum_{n=1}^{N}A_{n}^{T}L_{n}A_{n} &  -A_{1}^{T}L_{1}&  \cdots & -A_{N}^{T}L_{N}\\
    -L_{1}A_{1}&  L_{1}&  \cdots & 0\\
    \vdots &  \vdots&  \ddots & \vdots \\
    -L_{N}A_{N}&  0&  \cdots& L_{N}
\end{bmatrix}
}_{P}
\begin{bmatrix}
    Z\\
    R_{1}\\
    \vdots\\
R_{N}\nonumber
\end{bmatrix}.
\end{eqnarray}
According to \cite{bishop2006pattern}, matrix $P$ above is the precision matrix of the joint distribution, where the covariance matrix can be calculated as:
\begin{equation}
\label{eq:covjoint}
    \operatorname{cov}[Z, R_1, .., R_N]= P^{-1}.
\end{equation}
Next, the linear terms in Equation~\ref{eq15} can be decomposed as:
\begin{equation}
\label{eq:meanjoint}
Z^\top\Lambda \mu = \begin{bmatrix}
Z,
R_{1},
\cdots,
R_{N}
\end{bmatrix}\begin{bmatrix}
\Lambda \mu,
0,
\cdots,
0
\end{bmatrix}^\top.
\end{equation}
Then from~\cite{bishop2006pattern}, the mean of the joint distribution is computed by:
\begin{eqnarray}    
\mathbb{E}[Z, R_1, .., R_N] = P^{-1}\begin{bmatrix} 
    \Lambda \mu,
    0,
    \cdots,
0
\end{bmatrix}^\top
=\begin{bmatrix}
    \mu,
    A_{1} \mu,
    \cdots,
A_{N} \mu
\end{bmatrix}^\top.
\end{eqnarray}
The mean and covariance of the marginal distribution of $p(R_1, .., R_N)$ can be extracted from Equation \ref{eq:meanjoint} and \ref{eq:covjoint}:
\begin{equation}
\mathbb{E}\begin{bmatrix}
    R_{1}&\cdots  &R_{N}
  \end{bmatrix}=\begin{bmatrix}
   A_{1} \mu,
  \cdots,
  A_{N}\mu
  \end{bmatrix}^\top,
\end{equation}
 
\begin{equation}
\operatorname{cov}[R_{1} \cdots R_{N}] = \begin{bmatrix}
L_{1}^{-1} + A_{1}\Lambda^{-1} A_1^\top\\
\vdots\\
L_{N}^{-1} + A_{N}\Lambda^{-1} A_{N}^\top
\end{bmatrix}.
\end{equation}
Finally, with $p(Z, R_1, .., R_N)$, and $p(R_1, .., R_N)$, we could obtain the probability of $p(Z|R_1, .., R_N)$ through Gaussian conditioning, which is also a Gaussian:
\begin{equation}
    {\textstyle \sum_{Z|R_{1}\cdots R_{N}}}=\left(L + \sum_{n=1}^{N}A_{n}\Lambda A_{n}^\top\right)^{-1} ,
\end{equation}

\begin{eqnarray}    
\mu_{z|R_{1}\cdots R_{N}}= {\textstyle \sum_{z|R_{1}\cdots R_{N}}}\left( \Lambda \mu + \sum_{n=1}^{N}A_{n}^TL_{n}R_{n}  \right) .
\end{eqnarray}
In GBA, the Gaussian distributions assume to be factorized; thus $\Lambda^{-1}=\operatorname{diag}(\sigma_z)$ and $L_n^{-1}=\operatorname{diag}(\sigma_{R_n})$. 
The transformation $A_n=I(a_n)$, where $a_n$ represents the distance between a target node and the context node $n$. Then, the Gaussian can be further weight as:
\begin{align}
    &\bar{\sigma}_{Z}^{2}=\left[\left(\sigma_{Z}\right)^{-2}+\sum_{n=1}^{N}\left( \sigma_{R_n} / a_n\right)^{-2}\right]^{-1},\\
    &\bar{\mu}_{Z}=\bar{\sigma}_{Z}^{2} \left( \mu_{Z} / \sigma^2_{Z} + \sum_{n=1}^{N} a_n R_{n} / \sigma^2_{R_n} \right).
\end{align}

\section{Derivation of ELBO for STGNP}
\label{sec:elbo}
Here, we derive the evidence lower-bound (ELBO) for STGNP. For brevity, we still omit the subscript $m$. 
\begin{align}
\begin{split}
    &\log p\left(Y \mid X, \mathcal{C}, A\right)
    =\log \mathbb{E}_{q\left(Z \mid \mathcal{C}\cup \mathcal{D}, A\right)}\frac{p\left(Y, Z \mid X, \mathcal{C}, A\right)}{q\left(Z \mid \mathcal{C}\cup \mathcal{D}, A\right)}\\
    %&\geq \mathbb{E}_{q\left(Z \mid \mathcal{C}\cup \mathcal{D}, A\right)} \left[ \log \frac{p\left(Y, Z \mid X, \mathcal{C}, A\right)}{q\left(Z \mid \mathcal{C}\cup \mathcal{D}, A\right)} \right]\\
    &\geq \mathbb{E}_{q\left(\cdot\right)} \left[ \log \frac{
    p\left(Y \mid X, Z\right) p\left(Z \mid X, \mathcal{C}, A\right)}
    {q\left(Z \mid \mathcal{C}\cup \mathcal{D}, A\right)} \right]\\
    &= \mathbb{E}_{q\left(\cdot\right)} \left[ \log p\left(Y \mid X, Z\right) - \log \frac{q\left(Z \mid \mathcal{C}\cup \mathcal{D}, A\right)}{p\left(Z \mid X, \mathcal{C}, A\right)} \right]\\
    &= \mathbb{E}_{q\left(\cdot\right)} \left[ \log p\left(Y \mid X, Z\right) - \log \frac{\prod_{l=1}^{L} q\left(Z^l \mid Z^{l+1}, {V^\prime}^l, H^l, A\right)}{\prod_{l=1}^{L} q\left(Z^l \mid Z^{l+1}, V^l, H^l, A\right)} \right]\\
    &= \mathbb{E}_{q\left(Z \mid \mathcal{C}\cup \mathcal{D}, A\right)} \left[ \log p\left(Y \mid X, Z\right)\right] - \\
    &\sum_{l=1}^L \mathbb{E}_{q(Z^{l+1})} \left[\mathbb{KL}\left(q(Z^{l}|Z^{l+1}, {V^\prime}^l, H^l, A)||q(Z^l|Z^{l+1}, V^l, H^l, A)\right)\right].\\
\end{split}
\end{align}

\section{Experimental Details}
\subsection{STGNP Architectures}

\label{sec:arch}
\begin{figure}[!h]
  \centering
  \includegraphics[width=0.9\linewidth]{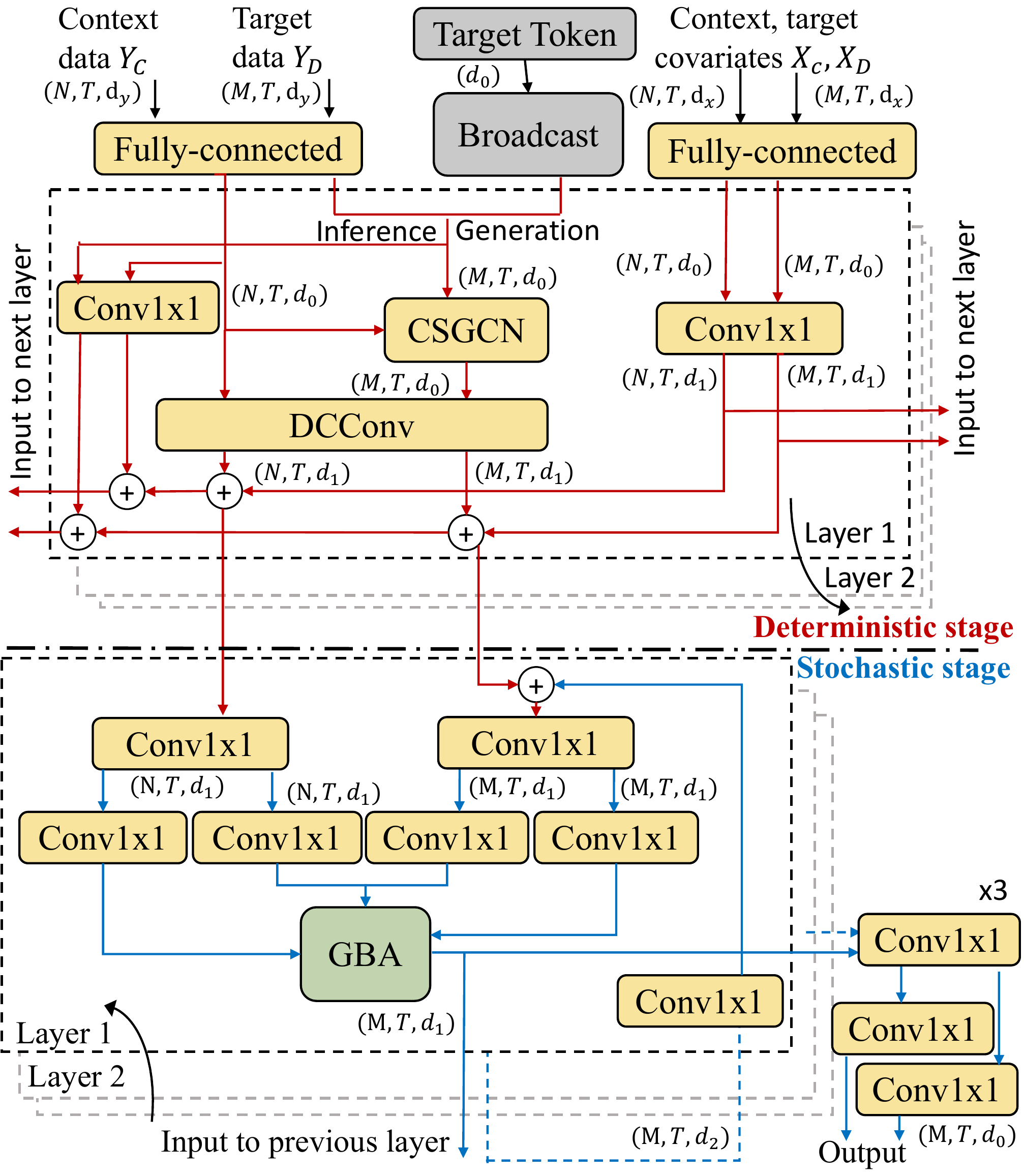}
  \caption{Detailed Architecture of STGNP.}
\label{fig:arch}
\end{figure}

Our STGNP has two major architectural components for deterministic and stochastic learning as shown in Figure~\ref{fig:arch}:
\begin{itemize}[leftmargin=*]
\item \textbf{Deterministic spatio-temporal stage:} The core component of this stage is the spatio-temporal learning module consisting of dilated causal convolutions and cross-set graph neural networks. Each layer contains a CSGCN to learn spatial dependencies and a DCconv to model temporal relations of sensor data.
A DCconv has a kernel size of $k=3$ and the number of output channels is denoted by $d$. A CSGCN includes a multilayer perceptron layer with $d$ neurons. 
We stack 3 layers with skip connections to capture spatial-temporal dependencies with $d=[16,32,64]$. The dilation factor is set to have an exponentially increasing rate of $2$ w.r.t the layers so the receptive field reaches $15$. Although this field is smaller than the length of the sequence ($T=24$), the ablation study shows that it still produces satisfactory performance.

\item \textbf{Stochastic generative stage:} The Graph Bayesian Aggregation and the likelihood function are major modules in this stage. The GBA includes a prior and a latent observation module. Both two modules contain a $1$-layer $1\times 1$ convolution with $d$ kernels followed by $2$ $1\times 1$ convolutions to obtain the mean and variance. 
We stack $3$ GBAs corresponding to the $3$ blocks in the first stage with $d=[16, 32, 64]$. The likelihood function generating extrapolation results is a $3$-layer $1\times 1$ convolution with $128$ channels. 
\end{itemize}

\subsection{Training Settings}
\label{sec:training}
All parameters are initialized with Xavier normalization~\cite{glorot2010understanding} and optimized by the Adam optimizer~\cite{kingma2014adam} with a learning rate of 10$^{-3}$. We train each model for $150$ epochs. At each iteration, we randomly sample $N-3$ nodes to extrapolate the remaining $3$ nodes, with the time length $T=24$. Note that the number of target nodes has an impact on the performance of the trained models. We conducted experiments to determine the optimal number of target nodes and found that using $3$ nodes generally resulted in the best performance across all baseline models.
%For model evaluation, we use all $N$ context nodes to extrapolate $M$ target nodes and we use rolling extrapolation to recover the first bunch of length $T=24$ and then the second, etc.

\subsection{Cross-Domain Evaluation}
We first learn models using the Beijing dataset and then evaluate their performances on the London dataset. As the London dataset lacks weather information, we remove this attribute in Beijing during training. The other training procedures remain the same. 
Note that we only investigate the performance of PM2.5 concentration and exclude stations BX1, and HR1 due to their large portion of missing values. This is because, for the London dataset, both PM10 and NO2 have a significant amount of missing data (5/7 stations without any signal, and 2/1 stations with missing rates larger than 60\%), which makes training unstable and the performance of the models largely depends on the training and testing data split.

\section{Additional Visualizations}
\label{sec:vis}
\begin{figure}[!b]
\centering
\subfigure{
\begin{minipage}[t]{1\linewidth}
\centering
\includegraphics[width=1\linewidth]{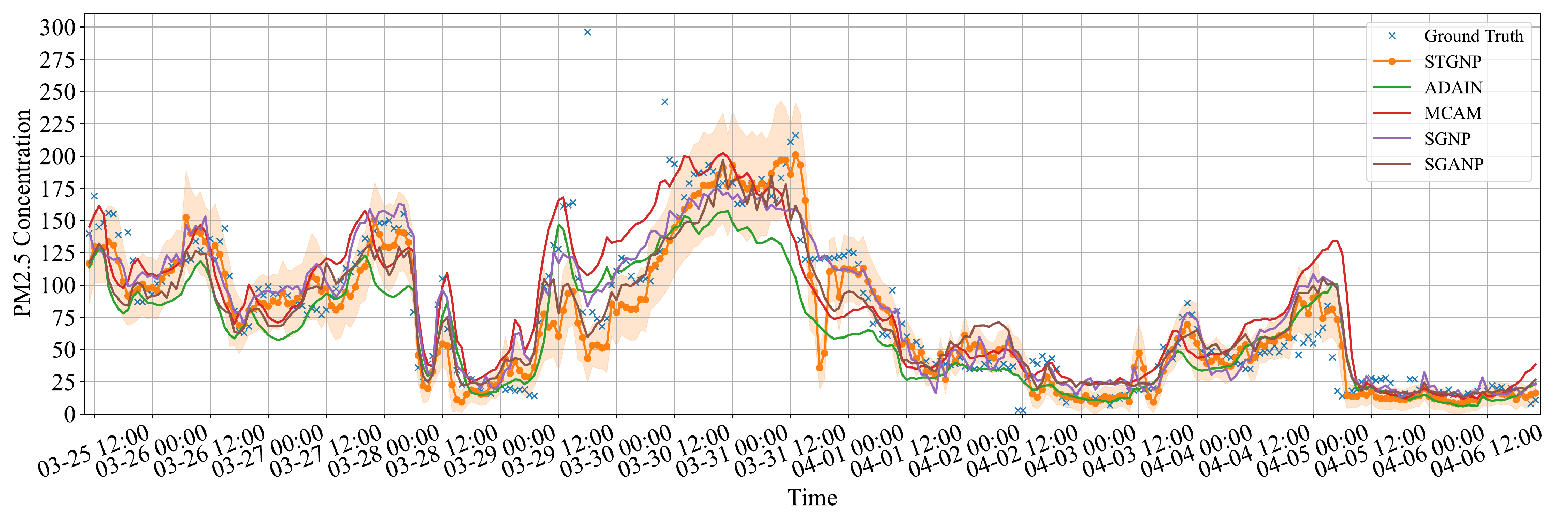}
%\caption{fig1}
\end{minipage}%
}

\subfigure{
\begin{minipage}[t]{1\linewidth}
\centering
\includegraphics[width=1\linewidth]{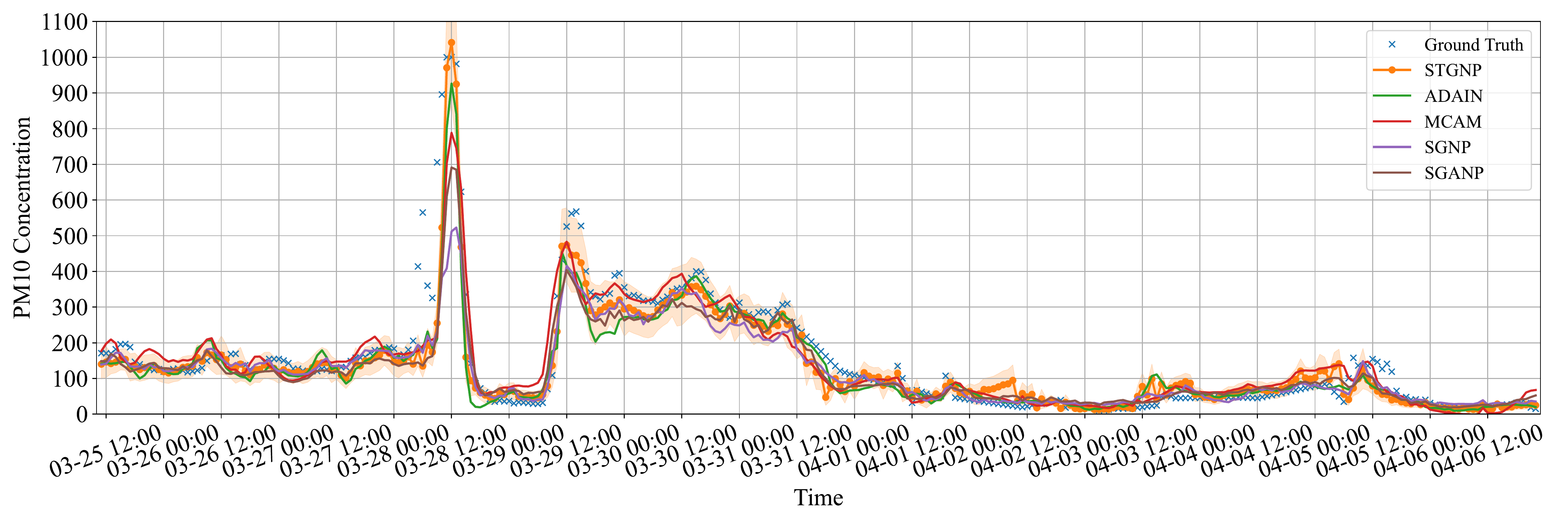}
% (b) Size of hidden states for ASGGRU.
%\caption{fig2}
\end{minipage}
}

\subfigure{
\begin{minipage}[t]{1\linewidth}
\centering
\includegraphics[width=1\linewidth]{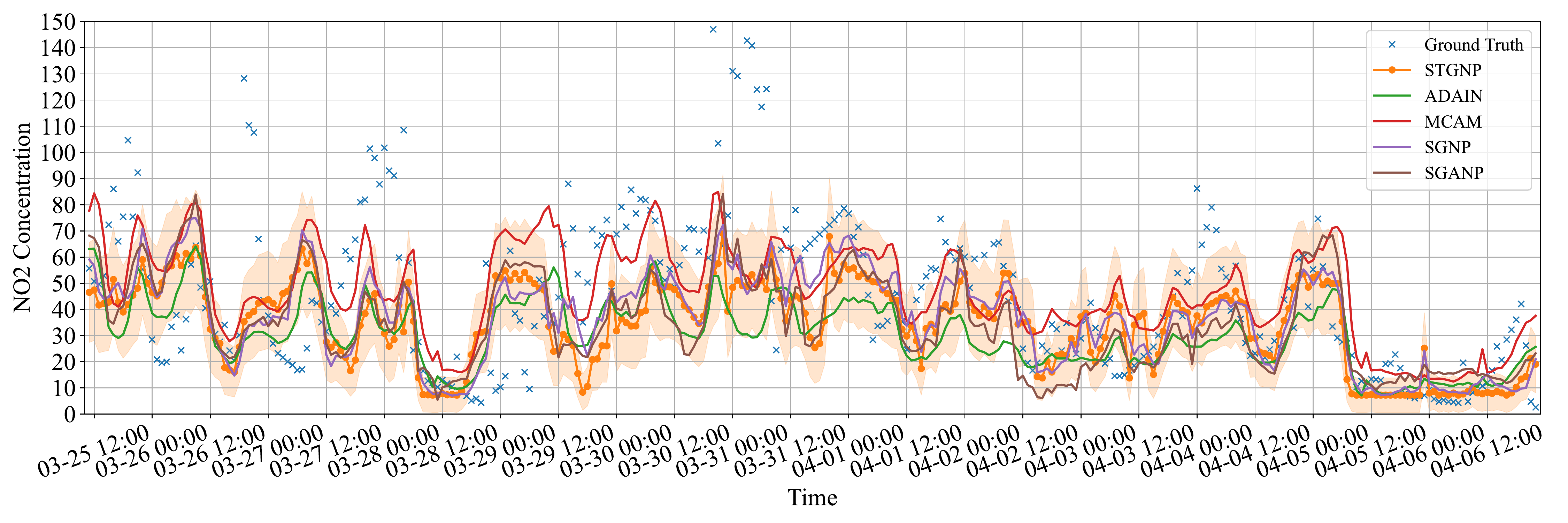}
% (c) Number of input sequence length $T$.
%\caption{fig2}
\end{minipage}
}%\hskip -10pt

\centering
\caption{Visualizations of PM2.5, PM10, and NO2 extrapolations of station 1019 on the Beijing dataset.}
\label{figure_visualization}
\end{figure}

We present additional visualization results of STGNP and the baselines on the Beijing dataset. As shown in Figure~\ref{figure_visualization}, our model consistently outperforms others on all time steps. Additionally, on less accurate extrapolations, our model is able to yield large uncertainties, which facilitates decision-making.

% \begin{figure}[b]
%   \centering
%   \includegraphics[width=1.\linewidth]{figure/appendix_visualization/BJ_PM25/Node_1015.pdf}
%   \caption{PM2.5 extrapolation performances of the station 1015.}
% \end{figure}

% \begin{figure}
%   \centering
%   \includegraphics[width=1\linewidth]{figure/appendix_visualization/BJ_PM10/Node_1019.pdf}
%   \caption{PM10 extrapolation performances of the station 1019.}
% \end{figure}

% \begin{figure}
%   \centering
%   \includegraphics[width=1\linewidth]{figure/appendix_visualization/BJ_NO2/Node_1019.pdf}
%   \caption{NO2 extrapolation performances of the station 1019.}
% \end{figure}

\end{document}